\documentclass[lettersize,twoside,journal]{IEEEtran}
\usepackage{amsmath,amsfonts}
\usepackage{algorithmic}
\usepackage{algorithm}
\usepackage{array}
\usepackage[caption=false,font=normalsize,labelfont=sf,textfont=sf]{subfig}
\usepackage{textcomp}
\usepackage{stfloats}
\usepackage{url}
\usepackage{verbatim}
\usepackage{graphicx}
\usepackage{cite}
\usepackage{booktabs} 
\usepackage{color}         % colors
\usepackage{xcolor}
\usepackage{colortbl}
\usepackage{adjustbox}
\usepackage{multirow}
\usepackage{makecell}
\usepackage[colorlinks=true, linkcolor=blue, citecolor=blue, urlcolor=black]{hyperref}
\usepackage[T1]{fontenc}
\definecolor{LightPurple}{rgb}{0.88,0.88,1}
\usepackage{enumitem}
\usepackage{diagbox}
\usepackage{pifont}
\usepackage{xr}
\usepackage{etoolbox}
\usepackage{orcidlink}
\begin{document}
	\title{Complementary Text-Guided Attention \\for Zero-Shot Adversarial Robustness}
	
	\author{Lu Yu\(^{\orcidlink{0000-0003-0578-6869}}\), Haiyang Zhang\(^{\orcidlink{0009-0005-5717-7004}}\), and Changsheng Xu\(^{\orcidlink{0000-0001-8343-9665}}\),~\IEEEmembership{Fellow,~IEEE}
	\vspace{-1.1em}
		
	\thanks{Received 7 July 2025; revised 8 February 2026; accepted 25 February 2026. (\textit{Corresponding author: Changsheng Xu.}) Recommended for acceptance by J.Verbeek. This work was supported by the Beijing Natural Science Foundation L252032, in part by the National Natural Science Foundation of China under Grants 62476199, 62376196, U23A20387, Tianjin Natural Science Foundation under Grant 24JCJQJC00190.}
	\thanks{\IEEEcompsocthanksitem Lu Yu and Haiyang Zhang are with School of Computer Science and Engineering, Tianjin University of Technology, 300384, China (e-mail: luyu@email.tjut.edu.cn, zshy@stud.tjut.edu.cn).
			\IEEEcompsocthanksitem Changsheng Xu is with State Key Laboratory of Multimodal Artificial Intelligence Systems, Institute of Automation, Chinese Academy of Sciences, Beijing, 100190, China, and also with the School of Artificial Intelligence, University of the Chinese Academy of Sciences, Beijing, 100049, China (e-mail: csxu@nlpr.ia.ac.cn).}
	\thanks{Our code is available at \href{https://github.com/zhyblue424/TGA-ZSR}{https://github.com/zhyblue424/TGA-ZSR}.}
	\thanks{Digital Object Identifier 10.1109/TPAMI.2026.3669252}}

	\markboth{IEEE TRANSACTIONS ON PATTERN ANALYSIS AND MACHINE INTELLIGENCE, 2026}
	{Yu \MakeLowercase{et al.}: Complementary Text-Guided Attention for Zero-Shot Adversarial Robustness}
	
	\IEEEpubid{0162-8828~\copyright~2026 IEEE}
	%\IEEEpubidadjcol
	\maketitle
	
	\begin{abstract}
		Due to the impressive zero-shot capabilities, pre-trained vision-language models (e.g., CLIP), have attracted widespread attention and adoption across various domains. Nonetheless, CLIP has been observed to be susceptible to adversarial examples. Through experimental analysis, we have observed a phenomenon wherein adversarial perturbations induce shifts in text-guided attention. Building upon this observation, we propose a simple yet effective strategy: \textit{Text-Guided Attention for Zero-Shot Robustness (TGA-ZSR)}. This framework incorporates two components: Local Attention Refinement Module and Global Attention Constraint Module. Our goal is to maintain the generalization of the CLIP model and enhance its adversarial robustness: The Local Attention Refinement Module aligns the text-guided attention obtained from the target model via adversarial examples with the text-guided attention acquired from the original model via clean examples. This alignment enhances the model's robustness. Additionally, the Global Attention Constraint Module acquires text-guided attention from both the target and original models using clean examples. Its objective is to maintain model performance on clean samples while enhancing overall robustness. However, we observe that the method occasionally focuses on irrelevant or spurious features, which can lead to suboptimal performance and undermine its robustness in certain scenarios. To overcome this limitation, we further propose a novel approach called \textit{Complementary Text-Guided Attention (Comp-TGA)}. This method integrates two types of foreground attention: attention guided by the class prompt and reversed attention driven by the non-class prompt. These complementary attention mechanisms allow the model to capture a more comprehensive and accurate representation of the foreground. The experiments validate that TGA-ZSR and Comp-TGA yield 9.58\% and 11.95\% improvements respectively, in zero-shot robust accuracy over the current state-of-the-art techniques across 16 datasets.
	\end{abstract}
	
	
	\section{Introduction}
	\IEEEPARstart{L}{arge-scale} pre-trained vision-language models (VLMs) have showcased remarkable success in artificial intelligence by seamlessly integrating visual and textual data to understand complex multimodal information, such as CLIP~\cite{CLIP}. Leveraging vast datasets and powerful architectures such as BERT~\cite{BERT} and its variants~\cite{MacBERT,roberta}, these models adeptly capture semantic relationships between images and texts, offering significant advantages across numerous applications. From image classification~\cite{clip-adapter,yu2024exploiting,tao2024class} and semantic segmentation~\cite{denseclip} to image captioning~\cite{clipcap} and vision question answering~\cite{clipvqa}, pre-trained VLMs revolutionize how machines perceive and interact with multimodal information. Their importance lies in their ability to learn rich representations from varied data streams, enabling zero-shot learning and transfer learning across domains and tasks. Thus, ensuring the reliability of large-scale models is crucial. However, these models are vulnerable to adversarial attacks, as many other networks as demonstrated by recent studies~\cite{TeCoA,PMG-AFT}, even slight perturbations to input data can result in misclassification or altered outputs. Such attacks pose a significant challenge, particularly in critical applications like autonomous vehicles~\cite{clipdriving}, medical diagnosis~\cite{clipmedical}, and maritime navigation~\cite{li2024novel}, where the consequences of erroneous decisions can be severe. As these large-scale models become increasingly prevalent in real-world applications, understanding and mitigating the risks posed by adversarial attacks is essential to maintain trust and reliability in AI systems.
	
	Adversarial training~\cite{AT2,AT3,AT1} has emerged as a crucial technique in enhancing the robustness of deep learning models against adversarial attacks. By augmenting training data with adversarial examples generated through perturbations of input data, models are forced to learn more robust decision boundaries, thereby improving their resilience to adversarial manipulation. Given the rising significance of large-scale VLMs in various applications, understanding their vulnerability to adversarial attacks is essential. While adversarial training presents practical challenges when applied to downstream tasks, especially with large-scale models. Firstly, adversarial training typically involves generating adversarial examples during each training iteration, which increases the computational overhead and may lead to overfitting on the training data. This phenomenon is exacerbated in large-scale models with vast parameter spaces, where fine-tuning becomes more susceptible to overfitting. Moreover, adversarial training may not adequately prepare models for all possible adversarial scenarios, potentially leaving them vulnerable to unknown data distributions encountered in real-world settings. Exploring zero-shot adversarial robustness in these models is particularly \IEEEpubidadjcol pertinent as it sheds light on their ability to generalize and perform reliably in unseen scenarios. Additionally, considering the multimodal nature of VLMs, the exploration of zero-shot adversarial robustness offers insights into the complex interactions between visual and textual modalities, paving the way for more robust and trustworthy multimodal AI systems.

	Text-guided Contrastive Adversarial Training (TeCoA) method~\cite{TeCoA} represents the pioneering effort in investigating the zero-shot adversarial robustness of large-scale VLMs. They aim to bolster CLIP's zero-shot generalization capacity against adversarial inputs. While their primary focus lies on enhancing accuracy in the face of adversarial samples, this improvement comes at the expense of decreased performance on clean data. Subsequent work by PMG-AFT~\cite{PMG-AFT} builds upon this by introducing a pre-trained model guided adversarial fine-tuning technique, further enhancing both generalizability and adversarial robustness. However, despite the advancements made by both studies in enhancing CLIP's zero-shot robustness, significant questions regarding the interpretability of adversarial attacks and the efficacy of adversarial training remain unanswered. Specifically, \textit{the mechanisms through which adversarial attacks influence network outputs and the reasons behind the effectiveness of adversarial training strategies remain elusive}. In our paper, we delve into the text-guided attention shift phenomenon to shed light on how adversarial attacks alter model outputs. Leveraging these insights, we propose a simple yet effective strategy, TGA-ZSR, aimed at enhancing the robustness of the CLIP model and preserving its performance on clean examples.
	
	However, we observe that the original text-guided attention occasionally focuses on irrelevant features, and the constraints derived from these inaccurate attentions may hinder the model's ability to enhance robustness. To address this limitation, we further propose Comp-TGA to obtain a more comprehensive attention mechanism. Specifically, we leverage both class prompts and non-class prompts to better distinguish between the foreground and background of an image. Class prompts (e.g. \textit{ ``This is a photo of a [class]''}) guide the model to emphasize the class-relevant features, ensuring that the attention remains focused on critical objects. In contrast, non-class prompts (e.g. \textit{ ``This is not a photo of a [class]''}) encourage the model to differentiate the background from the foreground. By integrating complementary attention sources and refining the alignment process, our method enhances the model's ability to focus on relevant features while filtering out noise and irrelevant details.
	
	Building on this, we design the Local Attention Refinement Module (LARM) to effectively filter out irrelevant or misleading information, thereby mitigating the influence of adversarial attacks that exploit weaknesses in the model's decision-making process. Furthermore, to preserve the model's generalization capability on clean images, we incorporate the Global Attention Constraint Module (GACM). This module ensures consistent performance on clean data while enhancing the model against adversarial disruptions. Specifically, LARM refines text-guided attention maps at the feature level by aligning the text-guided attention derived from adversarial examples in the target model with that obtained from clean examples in the original model, ensuring that the model focuses on semantically consistent and discriminative regions. In contrast, GACM imposes global, model-level constraints that enforce consistency in the overall attention behavior between the origin and target models on clean examples. Together, these complementary modules enable the model to not only resist adversarial perturbations but also preserve its recognition performance on unperturbed inputs.
	
	Our main contributions are summarized as follows:
	\begin{itemize}[leftmargin=*]
		\item 
		We introduce text-guided attention to enhance zero-shot robustness on vision-language models while maintaining performance on clean samples.
		\item We propose a complementary text-guided attention mechanism that combines two types of foreground attention: one guided by class prompts and the other by non-class prompts. This fusion enables the model to identify a more accurate and comprehensive region of interest.
		\item
		We improve the interpretability of adversarial attacks for zero-shot robustness on vision-language models through a text-guided attention mechanism.
		\item
		The experimental results show that TGA-ZSR and Comp-TGA surpass previous state-of-the-art methods, establishing a new benchmark in model zero-shot robust accuracy.
	\end{itemize}
	
	The initial version of this paper was accepted at the Thirty-Eighth Annual Conference on Neural Information Processing Systems~\cite{TGA-ZSR}. Building upon the initial conference version, this extended manuscript introduces several significant improvements and contributions: \textbf{(1) Methodological Advancements:} While the original TGA-ZSR framework leveraged text-guided attention to enhance adversarial robustness, we observed that such attention mechanisms can occasionally be inaccurate, leading to suboptimal performance. In this extended version, we propose complementary text-guided attention (Comp-TGA), a refined mechanism designed to yield a more comprehensive and accurate representation of the foreground. This enhancement further boosts the model's adversarial robustness. \textbf{(2) Expanded Related Work:} We have expanded the related work and experimental comparison sections to include recent and relevant approaches such as LAAT~\cite{LAAT} and TTC~\cite{TTC}.  Notably, TTC is a test-time adaptation method, and our comparisons with it further demonstrate the superior effectiveness and robustness of our approach. \textbf{(3) More Experiments:} Additional experiments have been conducted, including the effects of different prompts and comparisons with a broader set of methods.

	\section{Related Work}
	\subsection{Pre-trained Vision-language Models} 
	In recent years, advancements in computer vision\cite{vit,ResNet,swin} have primarily relied on training models with image-label pairs to recognize predefined object categories. However, these approaches often overlook the inherent semantic connections between textual descriptions and visual content. Motivated by the remarkable progress witnessed in natural language processing (NLP), exemplified by breakthroughs like Transformer~\cite{transformer}, BERT~\cite{BERT}, and GPT-3~\cite{gpt3}, researchers are increasingly drawn to the prospect of using textual data to enhance the capabilities of DNNs. These methodologies are referred to as VLMs~\cite{ALIGN,CLIP,VLM2} and one prominent approach is to directly learn the semantic similarity between images and corresponding textual descriptions through image-text pairs. By aligning the embeddings of these two modalities, models like CLIP~\cite{CLIP}, ALIGN~\cite{ALIGN}, BLIP~\cite{BLIP}, Visual-BERT~\cite{imagebert}, and ALBEF~\cite{ALBEF} aim to achieve superior performance across various tasks. CLIP~\cite{CLIP} leverages a vast dataset of 400 million image-text pairs sourced from the internet and employs contrastive loss to effectively align the embeddings of both modalities, thereby enhancing the model's capabilities. Experimental results underscore the significant performance gains achieved by incorporating textual information into the model, with zero-shot performance surpassing that of earlier deep neural network architectures. However, despite its impressive zero-shot accuracy, experiments~\cite{TeCoA,PMG-AFT} reveal vulnerabilities to adversarial examples, resulting in a notable decline in robustness.

	\subsection{Adversarial Robustness} 
	Deep neural networks have been found to be vulnerable to adversarial examples~\cite{IPNN,PGD,deepfool,yu2023adversarial}, which can fool DNNs to produce false outputs, rendering trained models unreliable. To bolster robustness against such adversarial attacks, various advanced methods have been proposed, including data augmentation~\cite{data,data:Augmax,data:ADOID,yu2023quality}, adversarial training~\cite{AT1,AT2,AT3,AT4}, progressive self-distillation~\cite{PSD}, randomization strategy~\cite{RPF,random}, and adversarial purification~\cite{AP,AP1,AP2,li2025adbm, li2025importance, li2023recognizing}. While these strategies aim to improve DNNs' adversarial robustness, they often come with increased complexity or limited generalizability. Adversarial training~\cite{AT1,AT2,AT3,AT4} stands out as one of the most widely used and effective approaches for improving adversarial robustness. It follows a min-max optimization framework, where the inner maximization step generates adversarial examples to maximize the model's loss, and the outer minimization step updates model parameters to minimize this worst-case loss, enhancing resistance to adversarial attacks. After the emergence of CLIP~\cite{CLIP}, many subsequent works~\cite{calip,yao2023visual,CBM} have utilized CLIP as a backbone, yet little attention has been given to studying its adversarial robustness. CLIP is shown to be susceptible to adversarial examples~\cite{TeCoA} as well, posing a significant threat to downstream tasks utilizing CLIP as a backbone. Hence, investigating the adversarial robustness of CLIP is crucial.

	\subsection{Zero-shot Adversarial Robustness for VLMs}
	The visual-language model, trained on both image and text data, serves as a foundational model for various tasks. However, it has shown vulnerability to adversarial examples~\cite{TeCoA,PMG-AFT}, and training from scratch is time-intensive. TeCoA~\cite{TeCoA} was the first to explore zero-shot adversarial robustness for VLMs, aiming to enhance CLIP's adversarial robustness by minimizing the cross-entropy loss between image logits and targets. While TeCoA solely utilizes cross-entropy loss, yielding only marginal performance improvements, PMG-AFT~\cite{PMG-AFT} extends this approach by minimizing the distance between features of adversarial examples and those of the pre-trained model.  FARE~\cite{FARE} primarily focuses on maintaining high clean accuracy while improving model robustness, achieving this by constraining the distance between the original and target model embeddings. LAAT~\cite{LAAT} addresses the suboptimal performance caused by the naive use of text encoders and enhances adversarial robustness through an expansion algorithm. TTC~\cite{TTC} leverages the image encoder to generate perturbations that effectively counteract adversarial examples during inference.  Since background cues play a critical role in model predictions~\cite{li2021rethinking}, we further observe that adversarial perturbations introduce significant shifts in attention maps compared with clean examples. Leveraging this insight, we enhance model robustness by constraining it with text-guided attention.

	\section{Methodology}\label{Methodology}
	
	\subsection{Preliminaries and Problem Setup}
	Following the previous works~\cite{TeCoA,PMG-AFT}, we choose the CLIP model as the pre-trained VLMs for the image classification task. Given an image-text pair $(\mathbf{x},\mathbf{t})$, where $\mathbf{x}$ represents an image and $\mathbf{t}$ represents a textual prompt, CLIP learns to encode both the image and the text into fixed-dimensional embeddings. Let $f(\mathbf{x})$ denote the embedding of the image $\mathbf{x}$ and $g(\mathbf{t})$ denote the embedding of the text prompt $\mathbf{t}$, $\mathbf{y}$ is the one-hot vector label. For training or fine-tuning on the downstream tasks, we use the cross-entropy loss, denoted as $L(\mathbf{x}, \mathbf{t}, \mathbf{y})$.
	\begin{equation}
		L(\mathbf{x},\mathbf{t},\mathbf{y})= -\mathbb{E}_{i,j}\bigg[\mathbf{y}_{ij} \mathrm{log} \frac{\mathrm{exp}(\mathrm{cos}(f(\mathbf{x})_{i},g(\mathbf{t})_{j}))/\tau)}{\sum_k \mathrm{exp}(\mathrm{cos}(f(\mathbf{x})_{i},g(\mathbf{t})_{k}))/\tau)}\bigg]\label{con:CE}
	\end{equation}
	where we set $\mathbf{y}_{ij} = 1$ if the image-text pair is positive, otherwise,  $\mathbf{y}_ {ij} = 0$. $\tau$ is the temperature parameter and $\mathrm{cos}$ indicates calculating the cosine similarity of the two embeddings.

	\noindent\textbf{Adversarial Attacks.} Adversarial attacks are a concerning phenomenon where small, often imperceptible perturbations are intentionally applied to input data with the aim of deceiving a model into producing incorrect outputs, which can be mathematically expressed as:
	\begin{equation}
		\begin{aligned}
			&\mathbf{x}_a = \mathbf{x} + \mathbf{\delta}, \\
			&\text{s.t.} \quad \mathcal{F}(\mathbf{x}_a) \neq \mathbf{y}, \quad \|\mathbf{\delta}\|_\infty \leq \varepsilon. 
		\end{aligned}
	\end{equation}
	
	Here, \(\mathcal{F}\) represents the classification model, \(\mathbf{\delta}\) represents the adversarial perturbation, \(\mathbf{y}\) is the ground truth of the original input, and \(\varepsilon\) defines the strength of adversarial perturbation.
	
	These perturbations are crafted with the goal of causing the model to misclassify or generate erroneous predictions while appearing indistinguishable to human observers. The Projected Gradient Descent (PGD)~\cite{PGD} method is an iterative approach for crafting adversarial examples. It starts with the original input data and then iteratively adjusts the data in the direction that maximizes the model's loss function while ensuring the perturbed data remains within a specified perturbation budget.  Mathematically, the PGD attack can be expressed as follows:
	\begin{equation}
		\mathbf{x}_{a+1}=\Pi_{\mathbf{x}+\mathbf{S}}(\mathbf{x}_a+\varepsilon \cdot \mathrm{sign}(\bigtriangledown_{\mathbf{x}_a} L(\mathbf{x}_a,\mathbf{t},\mathbf{y})))
	\end{equation}
	Here, $L$ represents the loss function, $\mathbf{x}$ denotes the original input data, $\varepsilon$ controls the magnitude of perturbation, and $\bigtriangledown_\mathbf{x} L$ represents the gradient of the loss function with respect to the input data. By adding or subtracting $\varepsilon$ times the sign of this gradient to the original input data, the PGD attack generates adversarial examples that lead to misclassification or incorrect predictions by the model. $\Pi_{\mathbf{x}+\mathbf{S}}$ makes the perturbed data remain within an $\varepsilon$-neighborhood of the original input, preventing the generated adversarial examples from straying too far. $\mathbf{S}$ is a set of allowed perturbations that formalizes the manipulative power of the adversary.
	
	\begin{figure*}[t]
		\centering
		\includegraphics[width=\linewidth]{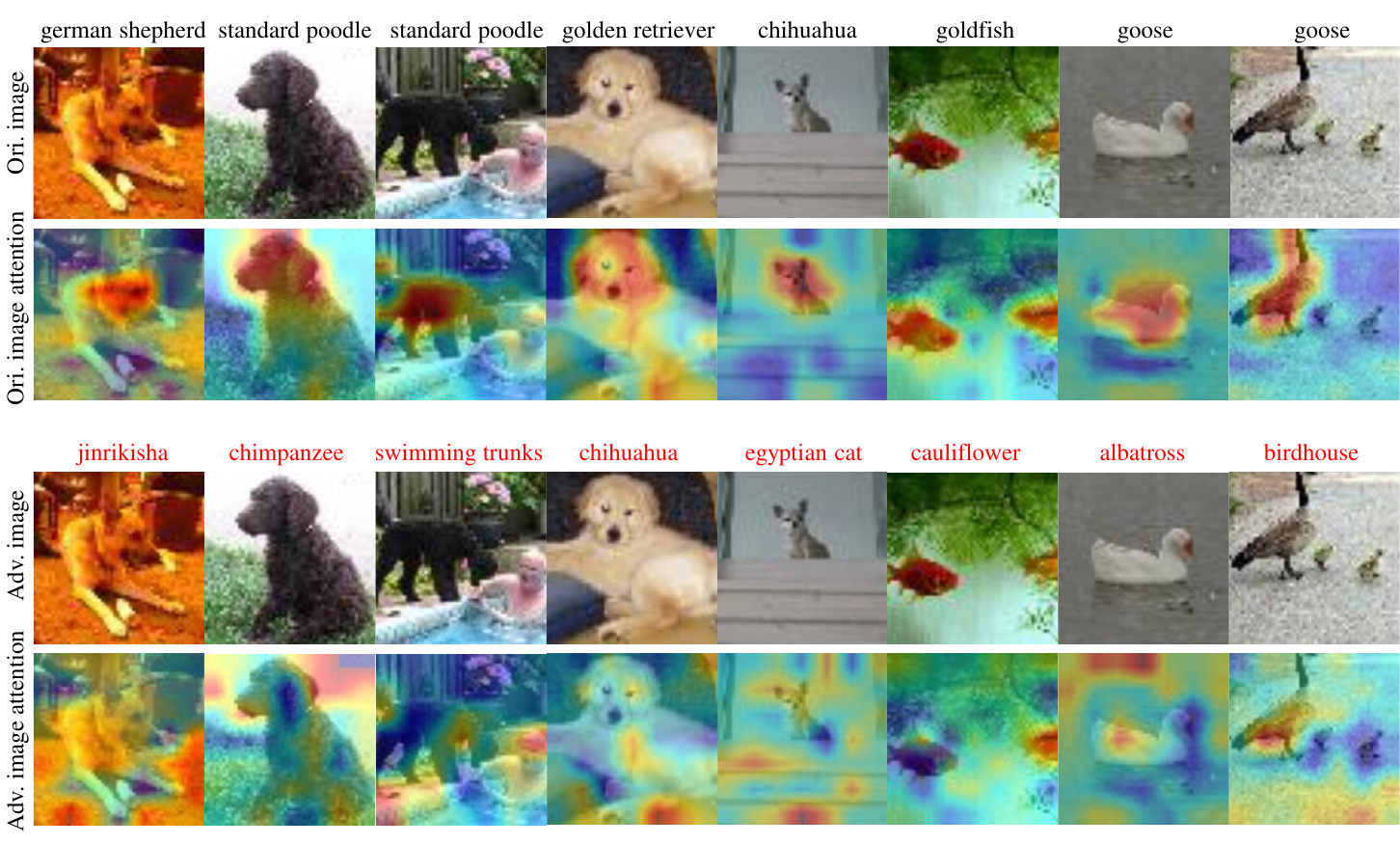}
		\caption{The four rows depict the original image, its associated attention map, the generated adversarial example, and the attention map of the adversarial example. Labels in black indicate the ground truth, while those in red represent misclassified labels for the adversarial examples.}
		\label{fig:interpretation}
	\end{figure*}
	
	\noindent\textbf{Adversarial Examples Generation and Adversarial Training.} The optimization objective for crafting adversarial examples aims to maximize the loss of model $f_{\mathbf{\theta}}$ with respect to a perturbed input $\mathbf{x}_a$, which can be formulated as:
	\begin{equation}
		\mathbf{x}_a = \mathop{argmax} \limits_{\mathbf{x}_a} L(f_{\mathbf{\theta}}(\mathbf{x}_a,\mathbf{t},\mathbf{y}))
	\end{equation}
	
	Adversarial training is a technique to generate adversarial examples from the original training data and then use these examples to train the model, forcing it to learn to resist adversarial perturbations. To adapt the model to the downstream tasks, we apply adversarial fine-tuning on one target model towards robustness with the following loss:
	\begin{equation}
		\mathbf{\theta} = \mathop{argmin} \limits_{\mathbf{\theta}}\mathcal{J}(f_{\mathbf{\theta}}(\mathbf{x}_a,\mathbf{t},\mathbf{y}))
	\end{equation}
	Where $\mathcal{J}$ represents the total loss function used for training the model.
	
	\noindent\textbf{Zero-Shot Adversarial Robustness.} In this paper, we investigate the zero-shot adversarial robustness of CLIP model, which refers to the ability of these models to maintain performance and reliability even when encountering unseen adversarial samples during inference, with only adversarial fine-tuning the original CLIP model on one target dataset, such as Tiny-ImageNet.

	\subsection{Text-Guided Attention based Interpretation of Adversarial Attacks}\label{sec:interpretation}
	\noindent\textbf{Text-Guided Attention.} Attention mechanisms~\cite{AM-ODD,calip,li2022exploring} play a crucial role in enhancing vision model performance across various tasks. At its core, attention enables models to focus on relevant parts of the input data while suppressing irrelevant information. Similarly, in VLMs, by incorporating textual guidance, the models can effectively focus on relevant visual features while processing language, thus facilitating more accurate and coherent multimodal understanding. Additionally, text-guided attention enhances interpretability by providing insights into the model's decision-making process, fostering trust and understanding in complex multimodal systems. Thus, we investigate the impact of text-guided attention on enhancing and interpreting zero-shot adversarial robustness in VLMs in this paper.
	We define the text-guided attention as follows:
	\begin{equation}
		A(x)=f_{g}(\mathbf{x}) \cdot g(\mathbf{t})^ \top , \quad A\in \mathbb{R}^{P \times 1}
		\label{attention}
	\end{equation}
	
	Where $f_{g}(\mathbf{x})$ represents the global image feature before the pooling operation of $f(\mathbf{x})$, and $P$ denotes the dimension of the attention embeddings. We reshape $A$ to $\mathbb{R}^{\sqrt{P}\times \sqrt{P}}$ to obtain the attention map, which is then resized to $A \in \mathbb{R}^{H\times W}$. Finally, we apply a normalization operation ($\mathrm{norm}$) on $A$ to obtain the final text-guided attention map.
	
	\begin{figure}[t]
		\centering
		\includegraphics[width=0.49\textwidth]{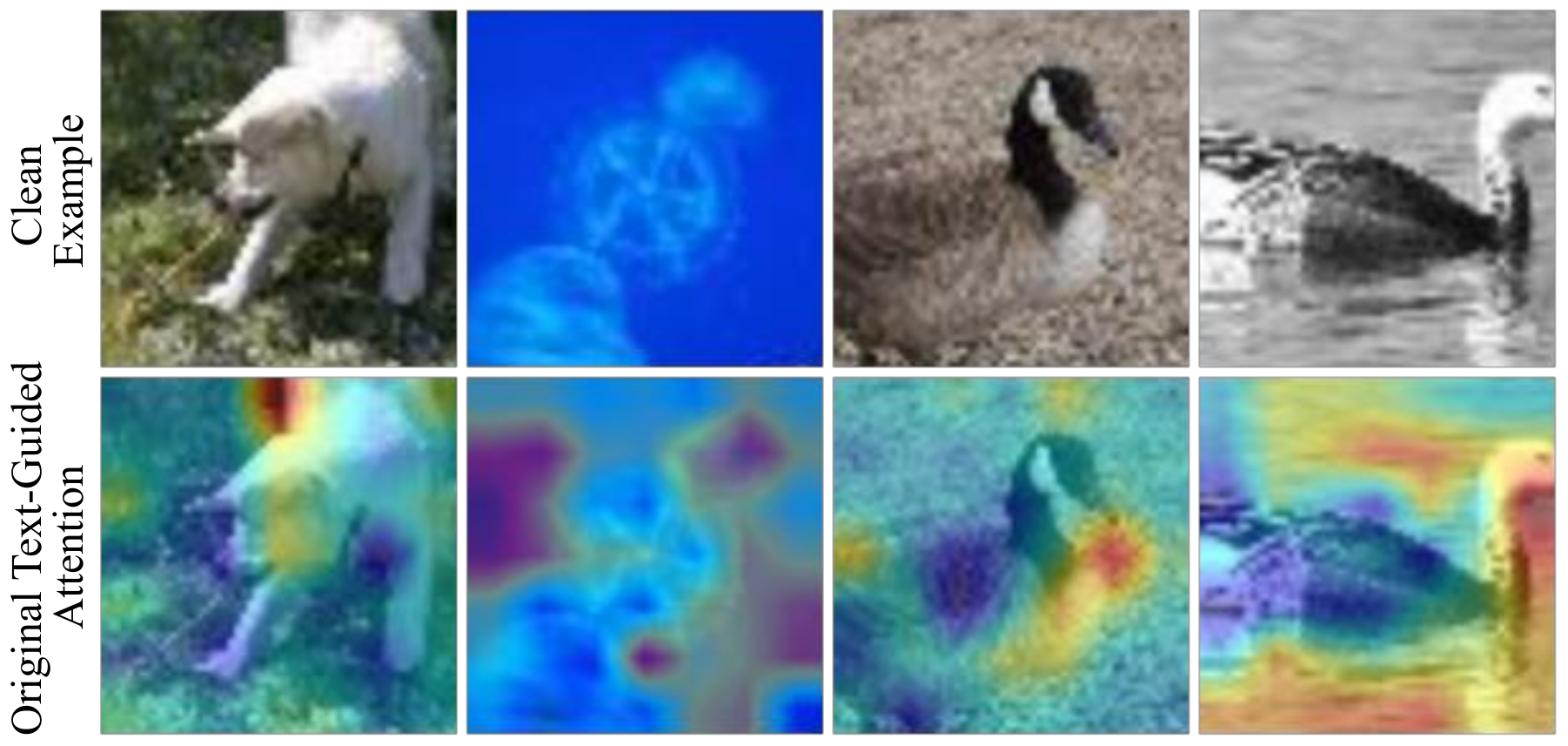}
		\caption{Examples of original text-guided attention for CLIP. These reveal that the attention mechanism occasionally focuses on irrelevant regions.}
		\label{atten}
	\end{figure}
	
	\begin{figure}[t]
		\centering
		\includegraphics[width=0.49\textwidth]{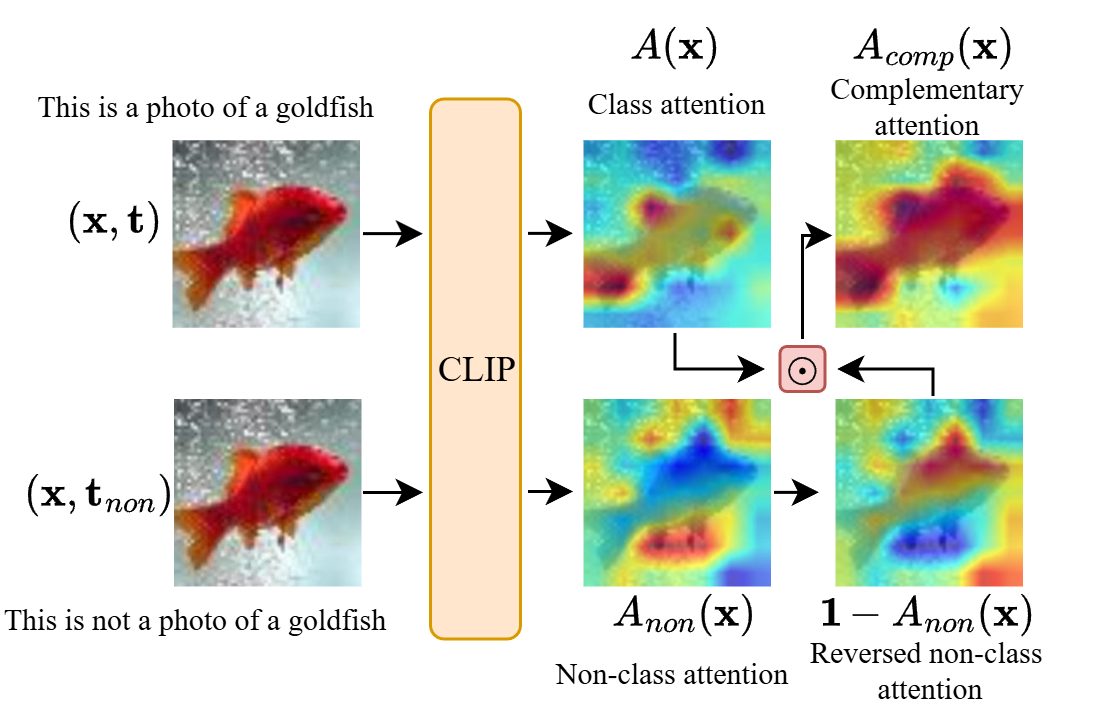}
		\caption{The framework overview of Complementary Text-Guided Attention.}
		\label{Comp-TGA}
	\end{figure}
	
	\noindent\textbf{Interpretation of Adversarial Attacks.}
	The previous research has predominantly focused on bolstering the zero-shot robustness of Vision-Language Models (VLMs), yet the reasons leading to misclassifications induced by adversarial attacks remain unclear. This paper aims to shed light on interpreting the impact of adversarial attacks on VLMs. By employing Eq.~\ref{attention}, we compute the text-guided attention for both the original image (\textit{Ori. image}) and its corresponding adversarial counterpart (\textit{Adv. image}), as depicted in Fig.~\ref{fig:interpretation}. Remarkably, despite the subtle discrepancies imperceptible to the human eye between the adversarial example and the original image, the former is misclassified (labels in red). However, a significant difference emerges in the respective text-guided attention maps. Specifically, we observe a notable shift in the text-guided attention of the adversarial example, characterized by instances of displacement towards other objects, backgrounds, or even disappearance. For instance, while the original images in the first, second, and fourth columns pay attention to their subjects' heads, in their adversarial counterparts, attention diverges elsewhere. In the third column, the attention shift leads from the correct object to an incorrect one, resulting in misclassification. In the fifth and seventh columns, the attention in their adversarial counterparts is redirected towards the background.

	\subsection{Complementary Text-Guided Attention based Interpretation of Adversarial Attacks}
	We further observe that the original text-guided attention occasionally focuses on the irrelevant features  (as shown in Fig.~\ref{atten}), and the constraints derived from these inaccurate attentions may hinder the model's ability to enhance robustness. To capture a more accurate representation of the foreground, we introduce a complementary fusion mechanism that integrates two sources of foreground information (as shown in Fig.~\ref{Comp-TGA}). By leveraging both direct foreground attention and an alternative foreground attention derived from background attention, it enhances the model's focus on the relevant features.

	\begin{figure}[t]
		\centering
		\includegraphics[width=0.49\textwidth]{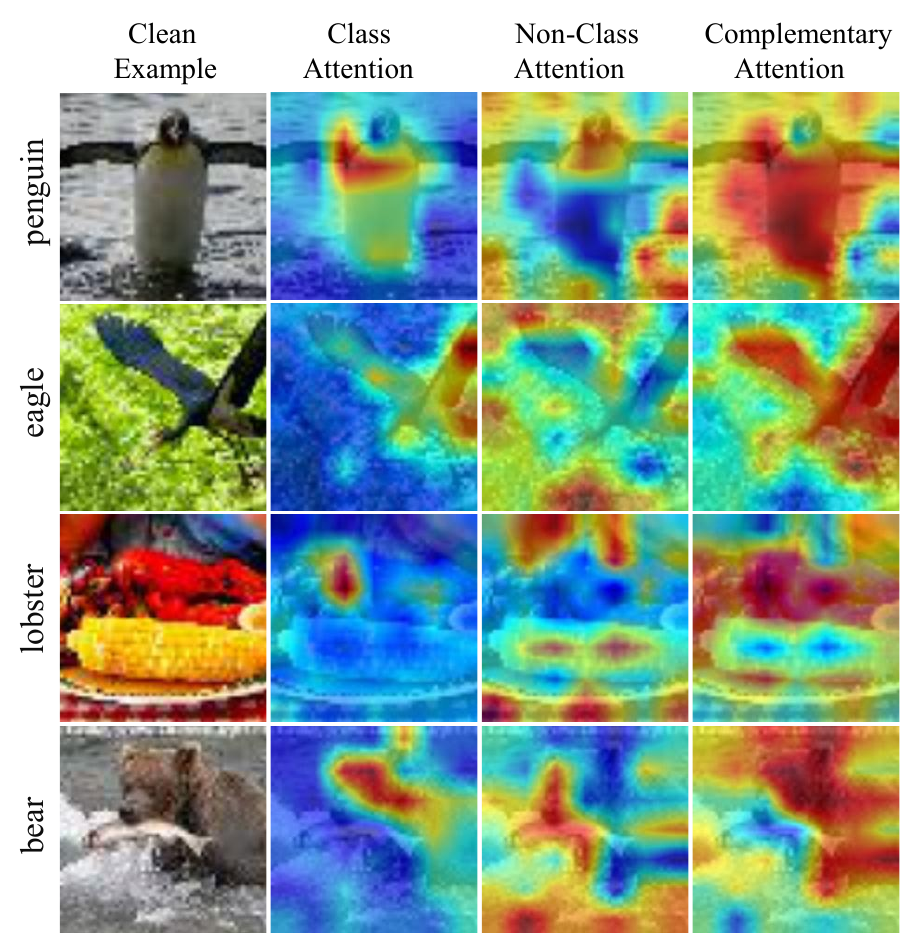}
		\caption{Visualization of different attention mechanisms. The complementary text-guided attention fusion generates a significantly more accurate and effective attention representation, addressing the limitations of the original text-guided attention.
		}
		\label{fig:visualization}
	\end{figure}

	First, we introduce class prompts (e.g. \textit{``This is a photo of a [class]''}) to guide the model's attention toward the foreground, where the class attention is obtained using text-guided attention, as defined in Eq.~\ref{attention}. To obtain the background of the examples, we introduce non-class prompts (e.g. \textit{``This is not a photo of a [class]''} for each image) to compute the non-class attention, defined as follows:
	\begin{equation}
		A_{non}(\mathbf{x}) = f_g(\mathbf{x}) \cdot g(\mathbf{t}_{non})^\top, \quad A_{non} \in \mathbb{R}^{P \times 1}
		\label{A_non}
	\end{equation}
	where \(\mathbf{t}_{non}\) represents the non-class prompts and \(g(\mathbf{t}_{non})\) denotes the text embeddings of \(\mathbf{t}_{non}\). Subsequently, an attention map in \(\mathbb{R}^{H \times W}\) is derived from \(A_{non} \in \mathbb{R}^{P \times 1}\) through a transformation process.
	
	Then, we reverse the non-class attention to obtain an alternative foreground representation, which is achieved by subtracting it from the all-one matrix \(\mathbf{1}\). The complementary attention fusion can be expressed as:
	\begin{equation}
		A_{comp}(\mathbf{x}) = A(\mathbf{x}) \odot (\mathbf{1}-A_{non}(\mathbf{x}))
		\label{fusion}
	\end{equation}
	where \(\odot\) denotes the Hadamard product, which is element-wise multiplication.
	
	This strategy enables the model to simultaneously capture two types of foreground attention, refining and ensuring the accuracy of the foreground representation. By distinguishing relevant features from irrelevant ones, it enhances the model's ability to focus on a more accurate foreground. The Fig.~\ref{fig:visualization} provides a comparative visualization of different attention mechanisms. The second column illustrates several failure cases of the original text-guided attention, which tends to drift away from the target object, revealing its limitations. The third column presents non-class attention, which identifies background regions, while the fourth column fuses both complementary attention mechanisms, effectively refining the focus on the target semantic regions. Notably, we provide visualizations across a diverse range of object categories, such as penguin, eagle, lobster, and bear, as well as complex multi-object images, including lobster \& corn and bear \& fish. The results demonstrate that the complementary attention mechanism consistently captures class-relevant information, even in challenging environments.

	\begin{figure*}[t]
		\centering
		\includegraphics[width=\linewidth]{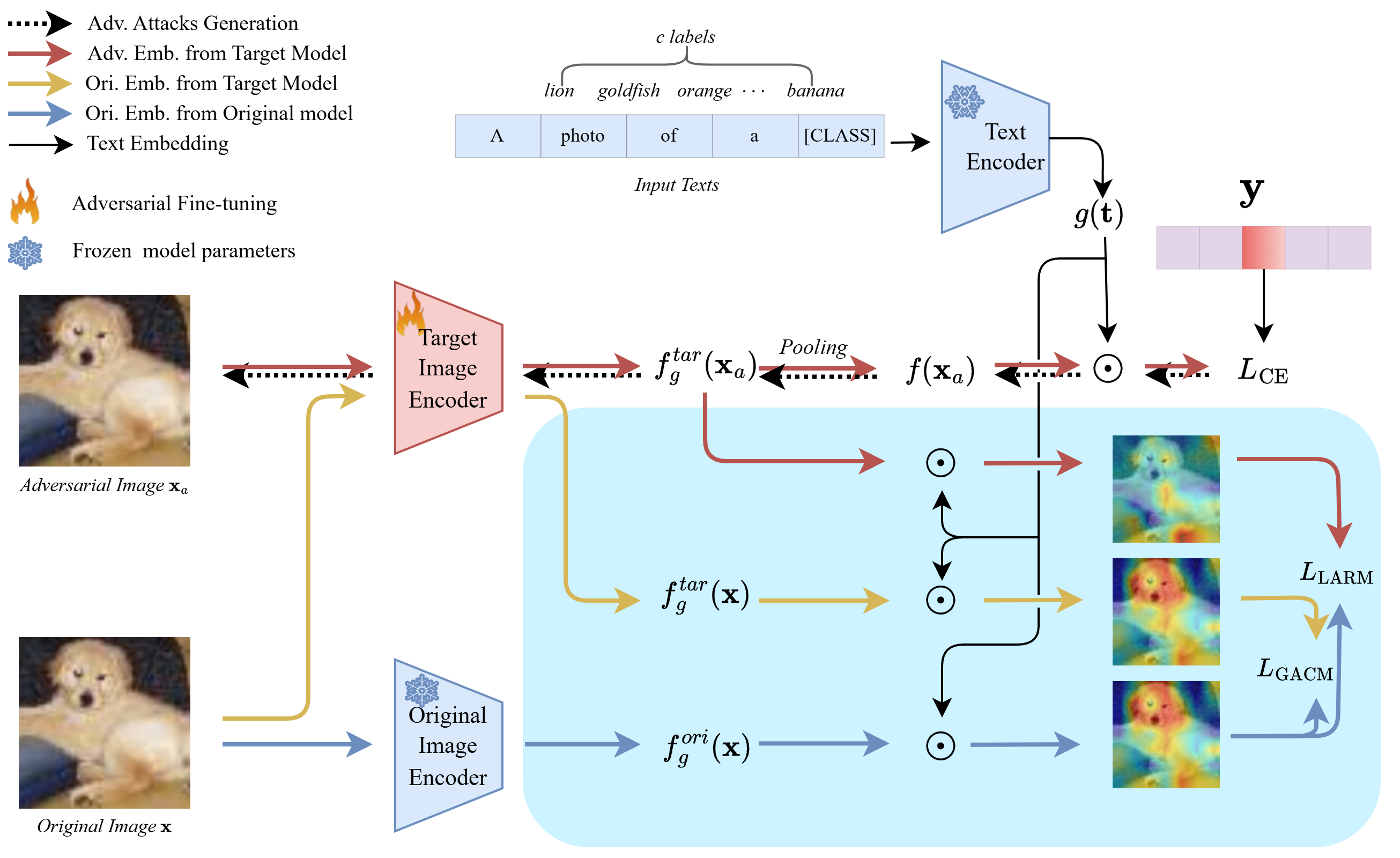}
		\caption{An overview of our TGA-ZSR framework: We generate adversarial examples and feed them into the target image encoder. To enhance the adversarial robustness of the CLIP model and maintain its generalization, we introduce text-guided attention. This involves refining the framework for adversarial examples through the Local Attention Refinement Module and constraining the model to prevent significant drift via the Global Attention Constraint Module.
		}
		\label{fig:frame}
	\end{figure*}

	\subsection{Model Training}
	The semantic information embedded within text representations is preserved through a frozen text encoder, offering invaluable guidance when adversarial perturbations disrupt relevant visual features, which has not been explored for zero-shot robustness of vision-language models. We introduce the Local Attention Refinement Module, designed to effectively filter out irrelevant information, thereby mitigating the impact of adversarial attacks seeking to exploit vulnerabilities in the model's decision-making process. Moreover, to maintain model's ability to generalize effectively on clean images, we introduce the Global Attention Constraint Module. This module ensures consistent performance on clean data while enhancing the model against adversarial disruptions. Additionally, employing text-guided attention enhances interpretability, offering crucial insights into how the model integrates and processes information across modalities. This interpretability not only instills trust in the model's predictions but also facilitates the detection and mitigation of adversarial attacks. Our approach \footnote{TGA-ZSR and Comp-TGA share the same framework, with the primary distinction being the approach used to obtain attention.} presents a comprehensive framework (as shown in Fig.~\ref{fig:frame}) for enhancing model robustness to adversarial perturbations while concurrently improving interpretability. We will introduce the details as follows.

	\noindent\textbf{Local Attention Refinement Module.}
	Based on the insights gained in Section~\ref{sec:interpretation}, we propose a local attention refinement module aimed at enhancing the robustness of the model. This module is designed to rectify the text-guided attention of adversarial samples, which often leads to altered predictions. Our approaches align the adversarial attention map with that of the clean samples, known for their high-accuracy attention distribution. This simple yet effective strategy serves to mitigate the impact of adversarial perturbations on the model's predictions. 
	
	We take the generated adversarial sample $\mathbf{x}_a$ to the target model $f_g^{tar}(\cdot)$ and the clean sample $\mathbf{x}$ to the original model $f_g^{ori}(\cdot)$ and obtain the adversarial attention map $\mathcal{A}(\mathbf{x}_a^i)_{tar}$ and the clean attention map $\mathcal{A}(\mathbf{x}^i)_{ori}$ respectively. The local attention refinement loss $L_{\textrm{LARM}}$ is thus defined as:
	\begin{equation}
		L_{\textrm{LARM}} = \frac{1}{N} \cdot \sum_{i=0}^{N} \| \mathcal{A}(\mathbf{x}_a^i)_{tar} - \mathcal{A}(\mathbf{x}^i)_{ori} \|_2 \\
		\label{L_ar}
	\end{equation}
	where $\| \|_2$ denotes the $L_2$ distance computation between two attention maps.
	
	In TGA-ZSR, the attention maps \footnote{We only compute the attention map for the image corresponding to the text prompt of the ground-truth label.} are defined as follows:
	\begin{equation}
		\begin{aligned}
			&\mathcal{A}(\mathbf{x}_a)_{tar}=f_{g}^{tar}(\mathbf{x}_a)\cdot g(\mathbf{t})^ \top \\
			&\mathcal{A}(\mathbf{x})_{ori}=f_{g}^{ori}(\mathbf{x})\cdot g(\mathbf{t})^ \top 
		\end{aligned}
	\end{equation}
	In contrast, the attention maps in Comp-TGA include an additional refinement step and are formulated as:
	\begin{equation}
		\begin{aligned}
			\mathcal{A}(\mathbf{x}_a)_{tar} &= A(\mathbf{x}_a)_{tar} \odot (\mathbf{1}-A_{non}(\mathbf{x}_a)_{tar}) \\
			& = (f_{g}^{tar}(\mathbf{x}_a)\cdot g(\mathbf{t})^ \top) \odot (\mathbf{1}- (f_{g}^{tar}(\mathbf{x}_a)\cdot g(\mathbf{t}_{non})^ \top))\\
			\mathcal{A}(\mathbf{x})_{ori} &= A(\mathbf{x})_{ori} \odot (\mathbf{1}-A_{non}(\mathbf{x})_{ori}) \\
			& = (f_{g}^{ori}(\mathbf{x})\cdot g(\mathbf{t})^ \top) \odot (\mathbf{1}- (f_{g}^{ori}(\mathbf{x})\cdot g(\mathbf{t}_{non})^ \top))\\
		\end{aligned}
	\end{equation}

	\noindent\textbf{Global Attention Constraint Module.}
	The Local Attention Refinement module serves to enhance the robustness of the models, consequently improving the accuracy of adversarial samples. However, this enhancement comes with a trade-off: it may marginally sacrifice the accuracy on clean samples due to shifts in model parameters. To preserve the generalization capability of pre-trained VLMs, we introduce a Global Attention Constraint Module. This module aims to mitigate performance drops on clean images, thereby ensuring the overall effectiveness and reliability of the model.
	
	Specifically, we input the clean sample $\mathbf{x}$ into the target model $f_g^{tar}(\cdot)$, adversarially fine-tuned on the Tiny-ImageNet dataset, to acquire the text-guided attention map $\mathcal{A}(\mathbf{x})_{tar}$. Concurrently, the original text-guided attention map outputted from the original CLIP model $f_g^{ori}(\cdot)$ is denoted as $\mathcal{A}(\mathbf{x})_{ori}$. To ensure the preservation of important parameters for clean images, we enforce an $L_2$ distance constraint between these two attention maps. The global attention constraint loss $L_{\textrm{GACM}}$ is formulated as:
	\begin{equation}
		L_{\textrm{GACM}}=\frac{1}{N} \cdot \sum_{i=0}^{N}  \left \| \mathcal{A}(\textbf{x}^i)_{tar}-\mathcal{A}(\textbf{x}^i)_{ori} \right \|_2 
		\label{L_amc}
	\end{equation}
	
	In TGA-ZSR, the attention maps are computed as:
	\begin{equation}
		\begin{aligned}
			&\mathcal{A}(\textbf{x})_{tar}=f_{g}^{tar}(\textbf{x})\cdot g(\textbf{t})^ \top \\
			&\mathcal{A}(\textbf{x})_{ori}=f_{g}^{ori}(\textbf{x})\cdot g(\textbf{t})^ \top 
		\end{aligned}
	\end{equation}
	while in Comp-TGA, the attention maps are given by:
	\begin{equation}
		\begin{aligned}
			\mathcal{A}(\textbf{x})_{tar} &= A(\textbf{x})_{tar} \odot (\mathbf{1}-A_{non}(\textbf{x})_{tar}) \\
			& = (f_{g}^{tar}(\textbf{x})\cdot g(\textbf{t})^ \top) \odot (\mathbf{1}- (f_{g}^{tar}(\textbf{x})\cdot g(\textbf{t}_{non})^ \top))\\
			\mathcal{A}(\textbf{x})_{ori} &= A(\textbf{x})_{ori} \odot (\mathbf{1}-A_{non}(\textbf{x})_{ori}) \\
			& = (f_{g}^{ori}(\textbf{x})\cdot g(\textbf{t})^ \top) \odot (\mathbf{1}- (f_{g}^{ori}(\textbf{x})\cdot g(\textbf{t}_{non})^ \top))\\
		\end{aligned}
	\end{equation}
	
	Thus, the final loss function, shared by both TGA-ZSR and Comp-TGA, can be represented as:
	\begin{equation}
		L_{\textrm{total}}=L_{\textrm{CE}}+\alpha \cdot L_{\textrm{LARM}} +\beta \cdot L_{\textrm{GACM}}
		\label{L_total}
	\end{equation}
	
	\begin{table*}[t]
\centering
\caption{Zero-shot robust accuracy on images attacked with $\varepsilon$ of 1/255 of PGD~\cite{PGD}. We performed several different methods on Tiny-ImageNet, using $\varepsilon$ of 1/255 of PGD to generate adversarial examples for training, and evaluated them across 16 datasets. 
The optimal accuracy is highlighted in \textbf{bold}, while the second-best accuracy is \underline{underlined}. The values in parentheses represent the standard deviation.}
\label{tab:main_result_adv}  
\begin{adjustbox}{width=\textwidth,keepaspectratio}
\begin{tabular}{c|c|ccccccccccccccc|c}

\toprule
& & \multicolumn{15}{c|}{\textbf{Zero-shot datasets}} & \\
\cline{3-17} 
\textbf{Methods} &\textbf{\makecell[c]{\small\rotatebox{75}{Tiny-ImageNet}}}  & \textbf{\makecell[c]{\small\rotatebox{75}{CIFAR-10}}} & \textbf{\makecell[c]{\small\rotatebox{75}{CIFAR-100}}} & \textbf{\makecell[c]{\small\rotatebox{75}{STL-10}}} & \textbf{\makecell[c]{\small\rotatebox{75}{SUN397}}} & \textbf{\makecell[c]{\small\rotatebox{75}{Food101}}} & \textbf{\makecell[c]{\small\rotatebox{75}{Oxfordpets}}} & \textbf{\makecell[c]{\small\rotatebox{75}{Flowers102}}} & \textbf{\makecell[c]{\small\rotatebox{75}{DTD}}}   & \textbf{\makecell[c]{\small\rotatebox{75}{EuroSAT}}} & \textbf{\makecell[c]{\small\rotatebox{75}{FGVC-Aircraft}}} & \textbf{\makecell[c]{\small\rotatebox{75}{ImageNet}}} & \textbf{\makecell[c]{\small\rotatebox{75}{Caltech-101}}} & \textbf{\makecell[c]{\small\rotatebox{75}{Caltech-256}}} & \textbf{\makecell[c]{\small\rotatebox{75}{StanfordCars}}} & \textbf{\makecell[c]{\small\rotatebox{75}{PCAM}}}  &\(\boldsymbol{A_{\textrm{Robust}}}\) \\

\midrule
CLIP~\cite{CLIP} &0.88 &2.42 &0.26 &26.11 &1.00 &6.60 &3.84 &1.19 &2.02 &0.05 &0.00 &1.24 &19.88 &12.60 &0.20 &0.11 &4.90 \\ 

FT-Clean &13.55 &19.92 &4.94 &40.00 &0.82 &0.64 &2.40 &0.68 &2.66 &0.05 &0.03 &1.08 &14.95 &9.69 &0.09 &1.32 &7.05 \\ 

FT-Adv. &51.59 &38.58 &21.28 &69.55 &17.60 &12.55 &34.97 &19.92 &15.90 &11.95 &1.83 &17.26 &50.73 &40.18 &8.42 &\textbf{48.88} &28.83 \\ 

TeCoA \cite{TeCoA}  &37.57 &30.30 &17.53 &67.19 &19.70 &14.76 &36.44 &22.46 &17.45 &12.14 &1.62 &18.18 &55.86 &41.88 &8.49 &47.39 &28.06 \\ 

FARE\cite{FARE} &23.88 &21.25 &10.72 &59.59 &8.30 &10.97 &24.56 &15.48 &10.96 &0.14 &0.84 &10.54 &45.96 &34.35 &4.38 &10.17 &18.25 \\  

LAAT\cite{LAAT} &52.71 &35.49 &21.32 &67.79 &18.51 &12.64 & 35.54 &19.29 &11.38 &12.60 &1.38 &18.43 &55.85 &41.23 &6.60 &37.47 & 28.02\\  

PMG-AFT\cite{PMG-AFT} &47.11 &46.01 &25.83 &74.51 &22.21 &19.58 &41.62 &23.45 &15.05 &12.54 &1.98 &21.43 &62.42 &45.99 &11.72 &\underline{48.64} &32.51 \\ 

\rowcolor{LightPurple} 
TGA-ZSR (ours) &\makecell[c]{\underline{63.95} \\ \scriptsize{($\pm$ 0.11)}} &\makecell[c]{\underline{61.45} \\ \scriptsize{($\pm$ 0.67)}} &\makecell[c]{\underline{35.27} \\ \scriptsize{($\pm$ 0.07)}} &\makecell[c]{\underline{84.22} \\ \scriptsize{($\pm$ 0.21)}}  &\makecell[c]{\underline{33.22} \\ \scriptsize{($\pm$ 0.39)}} &\makecell[c]{\textbf{33.97} \\ \scriptsize{($\pm$ 0.20)}}  &\makecell[c]{\underline{57.75} \\ \scriptsize{($\pm$ 0.76)}} &\makecell[c]{\underline{34.55} \\ \scriptsize{($\pm$ 0.35)}}  &\makecell[c]{\underline{22.08} \\ \scriptsize{($\pm$ 0.16)}} &\makecell[c]{\underline{14.27} \\ \scriptsize{($\pm$ 0.26)}}  &\makecell[c]{\underline{4.75} \\ \scriptsize{($\pm$ 0.27)}} &\makecell[c]{\underline{28.74} \\ \scriptsize{($\pm$ 0.11)}}  &\makecell[c]{\underline{70.97} \\ \scriptsize{($\pm$ 0.42)}} &\makecell[c]{\underline{60.06} \\ \scriptsize{($\pm$ 0.46)}}  &\makecell[c]{\underline{20.40} \\ \scriptsize{($\pm$ 0.68)}} &\makecell[c]{47.76 \\ \scriptsize{($\pm$ 0.35)}}  &\makebox[2pt][c]{\makecell[c]{\underline{42.09} \\ \scriptsize{($\pm$ 0.12)}}}  \\ 

\rowcolor{LightPurple}
Comp-TGA (ours) &\makecell[c]{\textbf{69.22} \\ \scriptsize{($\pm$ 0.84)}} &\makecell[c]{\textbf{70.46} \\ \scriptsize{($\pm$ 1.53)}} &\makecell[c]{\textbf{42.26} \\ \scriptsize{($\pm$ 1.21)}} &\makecell[c]{\textbf{86.23} \\ \scriptsize{($\pm$ 0.66)}}  &\makecell[c]{\textbf{34.61} \\ \scriptsize{($\pm$ 1.39)}} &\makecell[c]{\underline{33.57} \\ \scriptsize{($\pm$ 1.03)}}  &\makecell[c]{\textbf{60.09} \\ \scriptsize{($\pm$ 1.90)}} &\makecell[c]{\textbf{35.84} \\ \scriptsize{($\pm$ 1.34)}}  &\makecell[c]{\textbf{23.90} \\ \scriptsize{($\pm$ 0.63)}}  &\makecell[c]{\textbf{16.64} \\ \scriptsize{($\pm$ 0.88)}} &\makecell[c]{\textbf{5.41} \\ \scriptsize{($\pm$ 0.12)}}  &\makecell[c]{\textbf{29.47} \\ \scriptsize{($\pm$ 1.07)}} &\makecell[c]{\textbf{72.16} \\ \scriptsize{($\pm$ 0.47)}} &\makecell[c]{\textbf{60.83} \\ \scriptsize{($\pm$ 1.17)}}  &\makecell[c]{\textbf{22.17} \\ \scriptsize{($\pm$ 1.09)}} &\makecell[c]{48.56 \\ \scriptsize{($\pm$ 0.19)}}  &\makebox[2pt][c]{\makecell[c]{\textbf{44.46} \\ \scriptsize{($\pm$ 0.76)}}} \\ 

\bottomrule
\end{tabular}
\end{adjustbox}
% \vspace{-5mm}
\end{table*}
	\begin{table*}[t]
\centering
\caption{Zero-shot clean accuracy. We performed several different methods on Tiny-ImageNet, using $\varepsilon$ of 1/255 of PGD to generate adversarial examples for training, and evaluated them across 16 datasets.}
\label{tab:main_result_cle}  
\begin{adjustbox}{width=\textwidth,keepaspectratio}
\begin{tabular}{c|c|ccccccccccccccc|c}

\toprule
& & \multicolumn{15}{c|}{\textbf{Zero-shot datasets}} & \\
\cline{3-17} 
\textbf{Methods} &\textbf{\makecell[c]{\small\rotatebox{75}{Tiny-ImageNet}}}  & \textbf{\makecell[c]{\small\rotatebox{75}{CIFAR-10}}} & \textbf{\makecell[c]{\small\rotatebox{75}{CIFAR-100}}} & \textbf{\makecell[c]{\small\rotatebox{75}{STL-10}}} & \textbf{\makecell[c]{\small\rotatebox{75}{SUN397}}} & \textbf{\makecell[c]{\small\rotatebox{75}{Food101}}} & \textbf{\makecell[c]{\small\rotatebox{75}{Oxfordpets}}} & \textbf{\makecell[c]{\small\rotatebox{75}{Flowers102}}} & \textbf{\makecell[c]{\small\rotatebox{75}{DTD}}}   & \textbf{\makecell[c]{\small\rotatebox{75}{EuroSAT}}} & \textbf{\makecell[c]{\small\rotatebox{75}{FGVC-Aircraft}}} & \textbf{\makecell[c]{\small\rotatebox{75}{ImageNet}}} & \textbf{\makecell[c]{\small\rotatebox{75}{Caltech-101}}} & \textbf{\makecell[c]{\small\rotatebox{75}{Caltech-256}}} & \textbf{\makecell[c]{\small\rotatebox{75}{StanfordCars}}} & \textbf{\makecell[c]{\small\rotatebox{75}{PCAM}}}  &\(\boldsymbol{A_{\textrm{Clean}}}\) \\

\midrule
CLIP~\cite{CLIP} &57.26 &\textbf{88.06} &\underline{60.45} &\textbf{97.04} &\textbf{57.26} &\textbf{83.89} &\textbf{87.41} &\textbf{65.47} &\textbf{40.69} &\textbf{42.59} &\textbf{20.25} &\textbf{59.15} &\textbf{85.34} &\textbf{81.73} &\textbf{52.02} &\textbf{52.09} &\textbf{64.42} \\ 

FT-Clean &\textbf{79.04} &84.55 &54.25 &93.78 &46.80 &47.10 &80.98 &46.43 &30.32 &\underline{24.39} &9.30 &44.40 &78.69 &70.81 &31.15 &47.89 &54.37 \\ 

FT-Adv. &73.83 &68.96 &39.69 &86.89 &33.37 &27.74 &60.10 &33.45 &23.14 &16.49 &4.86 &32.06 &67.41 &57.72 &18.11 &49.91 &43.36 \\ 

TeCoA \cite{TeCoA}  &63.97 &66.14 &36.74 &87.24 &40.54 &35.11 &66.15 &38.75 &25.53 &17.13 &6.75 &37.09 &74.63 &62.50 &24.65 &50.01 &45.81 \\ 

FARE\cite{FARE} &\underline{77.54} &\underline{87.58} &\textbf{62.80} &\underline{94.33} &49.91 &\underline{70.02} &\underline{81.47} &\underline{57.10} &\underline{36.33} &22.69 &\underline{14.19} &\underline{51.78} &\underline{84.04} &\underline{77.50} &\underline{44.35} &46.07 &\underline{59.85} \\ 

LAAT\cite{LAAT} &73.96 &63.09 &39.32 &84.65 &35.62 &26.50 &58.84 &32.61 &19.68 &19.50 &3.78 &33.31 &72.66 &58.43 &16.01 &\underline{51.58} &43.10\\  

PMG-AFT\cite{PMG-AFT} &67.11 &74.62 &44.68 &88.85 &37.42 &37.47 &66.34 &35.66 &21.17 &17.76 &4.71 &35.93 &76.70 &61.96 &25.21 &49.99 &46.60 \\ 

\rowcolor{LightPurple}
TGA-ZSR (ours) &\makecell[c]{75.72 \\ \scriptsize{($\pm$ 0.12)}} &\makecell[c]{86.46 \\ \scriptsize{($\pm$ 0.26)}} 
&\makecell[c]{56.52 \\ \scriptsize{($\pm$ 0.35)}} &\makecell[c]{93.48 \\ \scriptsize{($\pm$ 0.19)}}  
&\makecell[c]{\underline{51.99} \\ \scriptsize{($\pm$ 0.25)}} &\makecell[c]{57.59 \\ \scriptsize{($\pm$ 0.34)}}  &\makecell[c]{77.32 \\ \scriptsize{($\pm$ 0.30)}} &\makecell[c]{48.08 \\ \scriptsize{($\pm$ 0.37)}} &\makecell[c]{29.06 \\ \scriptsize{($\pm$ 0.35)}} &\makecell[c]{24.24 \\ \scriptsize{($\pm$ 0.49)}}  &\makecell[c]{11.93 \\ \scriptsize{($\pm$ 0.27)}} &\makecell[c]{48.04 \\ \scriptsize{($\pm$ 0.06)}}  &\makecell[c]{80.70 \\ \scriptsize{($\pm$ 0.09)}} &\makecell[c]{74.74 \\ \scriptsize{($\pm$ 0.18)}}  &\makecell[c]{36.62 \\ \scriptsize{($\pm$ 1.03)}} &\makecell[c]{49.58 \\ \scriptsize{($\pm$ 0.17)}}   &\makebox[2pt][c]{\makecell[c]{56.44 \\ \scriptsize{($\pm$ 0.08)}}}  \\ 

\rowcolor{LightPurple}
Comp-TGA (ours) &\makecell[c]{76.07 \\ \scriptsize{($\pm$ 0.09)}} &\makecell[c]{84.46 \\ \scriptsize{($\pm$ 0.34)}} &\makecell[c]{55.64 \\ \scriptsize{($\pm$ 0.34)}} &\makecell[c]{92.88 \\ \scriptsize{($\pm$ 0.17)}}  &\makecell[c]{51.47 \\ \scriptsize{($\pm$ 0.09)}} &\makecell[c]{54.01 \\ \scriptsize{($\pm$ 0.46)}}  &\makecell[c]{76.49 \\ \scriptsize{($\pm$ 0.63)}} &\makecell[c]{46.07 \\ \scriptsize{($\pm$ 0.54)}}  &\makecell[c]{29.43 \\ \scriptsize{($\pm$ 0.54)}}  &\makecell[c]{23.25 \\ \scriptsize{($\pm$ 0.59)}} &\makecell[c]{11.30 \\ \scriptsize{($\pm$ 0.56)}}  &\makecell[c]{46.08 \\ \scriptsize{($\pm$ 0.09)}} &\makecell[c]{80.50 \\ \scriptsize{($\pm$ 0.60)}}  &\makecell[c]{73.31 \\ \scriptsize{($\pm$ 0.22)}} &\makecell[c]{36.55 \\ \scriptsize{($\pm$ 0.36)}}  &\makecell[c]{49.46 \\ \scriptsize{($\pm$ 0.12)}} &\makebox[2pt][c]{\makecell[c]{55.44 \\ \scriptsize{($\pm$ 0.22)}}} \\ 

\bottomrule
\end{tabular}
\end{adjustbox}
% \vspace{-5mm}
\end{table*}
	
	\section{Experiments}
	\subsection{Experimental Setup}
	\noindent\textbf{Datasets.} 
	Our experiments begin with training the pre-trained CLIP model on the Tiny-ImageNet~\cite{ImageNet}. Then we evaluate the model's zero-shot adversarial robustness across 15 subsequent datasets, followed by previous studies, such as TeCoA~\cite{TeCoA} and PMG-AFT~\cite{PMG-AFT}. These datasets include several commonly used classfication datasets, including CIFAR-10~\cite{CIFAR10/100}, CIFAR-100~\cite{CIFAR10/100}, STL-10~\cite{STL10}, ImageNet~\cite{ImageNet}, Caltech-101~\cite{Caltech101}, and Caltech-256~\cite{Caltech256}. Additionally, fine-grained image classification datasets such as StanfordCars~\cite{StanfordCars}, Flowers102~\cite{flowers102}, Food101~\cite{Food101}, FGVCAircraft~\cite{FGVCAircraft}, and OxfordPets~\cite{OxfordPets} are included. Furthermore, the scene recognition dataset SUN397~\cite{SUN397}, the medical image dataset PCAM~\cite{pcam}, the satellite image classification dataset EuroSAT~\cite{Eurosat} and the texture recognition dataset DTD~\cite{DTD} are incorporated for comprehensive evaluation.

	\noindent\textbf{Implementation Details.} \label{Implementation}
	Following the protocol of previous works~\cite{PMG-AFT}, we fine-tuned the CLIP model on the adversarial samples of Tiny-ImageNet~\cite{ImageNet} as \emph{`adversarial fine-tuning'} and subsequently evaluated its performance across 15 datasets and Tiny-ImageNet itself. We employ ViT-B/32 as the backbone in CLIP and utilize the SGD optimizer to minimize loss. During adversarial fine-tuning, we update all parameters of the image encoder with a learning rate of 1e-4, weight decay of 0, momentum of 0.9, and a batch size of 128. We utilize $l_\infty$ norm PGD-2~\cite{PGD} with 2 iterations to generate adversarial examples, with an attack strength $\varepsilon$ of 1/255 and the attack step size is 1/255. To evaluate zero-shot adversarial inference, we employ $l_\infty$ norm PGD-100~\cite{PGD} with 100 iterations, attack step of 1/255, and a batch size of 256 to generate adversarial examples for verifying CLIP's adversarial robustness. Additionally, to assess the model's robustness under different attack strengths, we perform inference using adversarial strengths $\varepsilon$ of 1/255, 2/255,  and 4/255. The hyperparameters $\alpha$ and $\beta$ are set to 0.08 and 0.05 respectively in Eq.~\ref{L_total} in the main experiments of TGA-ZSR, and the same values are used for the CW attack. For the AutoAttack~\cite{AutoAttack} experiments, $\alpha$ and $\beta$ are set to 0.08 and 0.009. For Comp-TGA, we adopt a similar configuration, setting \(\alpha\) and \(\beta\) to 0.10 and 0.10 for both the PGD and CW attacks, and adjusting them to 0.10 and 0.07 for the AutoAttack setting. We conducted the experiment utilizing the RTX 3090, which required a training period ranging from 3 to 4 hours. 
	
	\noindent\textbf{Evaluation Metrics.}
	Consistent with previous work, we report the average Robustness Accuracy \(A_{\textrm{Robust}}\) and Clean Accuracy \(A_{\textrm{Clean}}\) across all datasets to assess the model's performance on adversarial and clean examples, respectively. To further capture the trade-off between robustness and generalization exhibited by different methods, we further report the metric \(A_{\textrm{Overall}}\), which reflects the overall performance by averaging \(A_{\textrm{Robust}}\) and \(A_{\textrm{Clean}}\).
	
	\begin{table*}[t]
\centering
\caption{Zero-shot robust accuracy on images attacked with $\varepsilon$ of 1/255 of AutoAttack~\cite{AutoAttack}. We performed several different methods on Tiny-ImageNet, using $\varepsilon$ of 1/255 of PGD to generate adversarial examples for training, and evaluated them across 16 datasets.}
\label{tab:autoattack}  
\begin{adjustbox}{width=\textwidth,keepaspectratio}
\begin{tabular}{c|c|ccccccccccccccc|c}

\toprule
& & \multicolumn{15}{c|}{\textbf{Zero-shot datasets}} & \\
\cline{3-17} 
\textbf{Methods} &\textbf{\makecell[c]{\small\rotatebox{75}{Tiny-ImageNet}}}  & \textbf{\makecell[c]{\small\rotatebox{75}{CIFAR-10}}} & \textbf{\makecell[c]{\small\rotatebox{75}{CIFAR-100}}} & \textbf{\makecell[c]{\small\rotatebox{75}{STL-10}}} & \textbf{\makecell[c]{\small\rotatebox{75}{SUN397}}} & \textbf{\makecell[c]{\small\rotatebox{75}{Food101}}} & \textbf{\makecell[c]{\small\rotatebox{75}{Oxfordpets}}} & \textbf{\makecell[c]{\small\rotatebox{75}{Flowers102}}} & \textbf{\makecell[c]{\small\rotatebox{75}{DTD}}}   & \textbf{\makecell[c]{\small\rotatebox{75}{EuroSAT}}} & \textbf{\makecell[c]{\small\rotatebox{75}{FGVC-Aircraft}}} & \textbf{\makecell[c]{\small\rotatebox{75}{ImageNet}}} & \textbf{\makecell[c]{\small\rotatebox{75}{Caltech-101}}} & \textbf{\makecell[c]{\small\rotatebox{75}{Caltech-256}}} & \textbf{\makecell[c]{\small\rotatebox{75}{StanfordCars}}} & \textbf{\makecell[c]{\small\rotatebox{75}{PCAM}}}  &\(\boldsymbol{A_{\textrm{Robust}}}\) \\

\midrule
CLIP~\cite{CLIP} &0.02 &0.01 &0.08 &0.03 &0.04 &0.01 &0.00 &0.03 &0.16 &0.12 &0.06 &0.04 &0.43 &0.10 &0.11 &0.22 &0.09 \\ 

FT-Clean &0.08 &0.03 &0.01 &0.91 &0.09 &0.04 &0.06 &0.03 &0.48 &0.02 &0.03 &0.12 &1.38 &0.66 &0.03 &0.03 &0.25  \\ 

FT-Adv. &\underline{50.48} &37.55 &20.39 &69.14 &16.25 &11.23 &33.91 &18.54 &\textbf{19.95} &11.59 &1.65 &16.21 &49.90 &39.24 &7.57 &\textbf{48.84} &28.28 \\ 

TeCoA~\cite{TeCoA} &35.03 &28.18 &16.09 &66.08 &17.41 &13.05 &34.81 &20.80 &15.37 &11.40 &1.32 &16.32 &54.54 &40.15 &7.15 &47.12 &26.55 \\ 

FARE~\cite{FARE} &28.59 &23.37 &13.58 &60.70 &9.72 &13.88 &27.72 &15.48 &9.15 &0.25 &0.87 &12.07 &47.45 &36.68 &6.77 &10.23 &19.78 \\ 

LAAT~\cite{LAAT} &51.88 &34.82 &20.99 &67.41 &17.59 &12.03 &34.97 &18.38 &11.06 &\underline{11.99} &1.35 &17.52 &55.19 &40.75 &6.11 &37.47 &27.47\\

PMG-AFT~\cite{PMG-AFT} &44.26 &\textbf{44.12} &\underline{23.66} &\textbf{73.90} &19.63 &\textbf{17.25} &39.25 &20.87 &13.72 &\underline{11.99} &1.68 &19.17 &\textbf{60.57} &44.25 &9.59 &\underline{48.53} &\underline{30.78} \\

\rowcolor{LightPurple}
TGA-ZSR (ours) &\makecell[c]{49.49 \\ \scriptsize{($\pm$ 0.09)}} &\makecell[c]{40.69 \\ \scriptsize{($\pm$ 0.14)}} &\makecell[c]{22.30 \\ \scriptsize{($\pm$ 0.08)}} &\makecell[c]{71.97 \\ \scriptsize{($\pm$ 0.08)}}  &\makecell[c]{\underline{20.14} \\ \scriptsize{($\pm$ 0.19)}} &\makecell[c]{\underline{15.41} \\ \scriptsize{($\pm$ 0.15)}} &\makecell[c]{\underline{39.82} \\ \scriptsize{($\pm$ 0.57)}}  &\makecell[c]{\underline{21.35} \\ \scriptsize{($\pm$ 0.10)}} &\makecell[c]{16.81 \\ \scriptsize{($\pm$ 0.30)}}  &\makecell[c]{11.31 \\ \scriptsize{($\pm$ 0.10)}}  &\makecell[c]{\underline{2.34} \\ \scriptsize{($\pm$ 0.30)}} &\makecell[c]{\underline{19.22} \\ \scriptsize{($\pm$ 0.06)}}  &\makecell[c]{\underline{57.65} \\ \scriptsize{($\pm$ 0.50)}} &\makecell[c]{\textbf{45.73} \\ \scriptsize{($\pm$ 0.13)}}  &\makecell[c]{\underline{10.14} \\ \scriptsize{($\pm$ 0.29)}} &\makecell[c]{48.01 \\ \scriptsize{($\pm$ 0.60)}} &\makebox[2pt][c]{\makecell[c]{30.77 \\ \scriptsize{($\pm$ 0.07)}}} \\

\rowcolor{LightPurple}
Comp-TGA (ours) &\makecell[c]{\textbf{53.00} \\ \scriptsize{($\pm$ 0.04)}} &\makecell[c]{\underline{42.18} \\ \scriptsize{($\pm$ 0.01)}} &\makecell[c]{\textbf{23.58} \\ \scriptsize{($\pm$ 0.06)}} &\makecell[c]{\underline{73.04} \\ \scriptsize{($\pm$ 0.59)}}  &\makecell[c]{\textbf{20.47} \\ \scriptsize{($\pm$ 0.09)}} &\makecell[c]{15.30 \\ \scriptsize{($\pm$ 0.02)}}  &\makecell[c]{\textbf{40.28} \\ \scriptsize{($\pm$ 0.18)}} &\makecell[c]{\textbf{22.37} \\ \scriptsize{($\pm$ 0.04)}}  &\makecell[c]{\underline{17.17} \\ \scriptsize{($\pm$ 0.21)}}  &\makecell[c]{\textbf{12.28} \\ \scriptsize{($\pm$ 0.90)}} &\makecell[c]{\textbf{3.43} \\ \scriptsize{($\pm$ 0.03)}}  &\makecell[c]{\textbf{19.47} \\ \scriptsize{($\pm$ 0.07)}} &\makecell[c]{57.58 \\ \scriptsize{($\pm$ 0.39)}}  &\makecell[c]{\underline{45.11} \\ \scriptsize{($\pm$ 0.30)}} &\makecell[c]{\textbf{10.24} \\ \scriptsize{($\pm$ 0.02)}}  &\makecell[c]{47.05 \\ \scriptsize{($\pm$ 0.58)}} &\makebox[2pt][c]{\makecell[c]{\textbf{31.41} \\ \scriptsize{($\pm$ 0.10)}}} \\

\bottomrule
\end{tabular}
\end{adjustbox}
% \vspace{-5mm}
\end{table*}
	\begin{table*}[t]
\centering
\caption{Zero-shot robust accuracy on images attacked with $\varepsilon$ of 1/255 of CW~\cite{cw}. We performed several different methods on Tiny-ImageNet, using $\varepsilon$ of 1/255 of PGD to generate adversarial examples for training, and evaluated them across 16 datasets.}
\label{tab:cw}  
\begin{adjustbox}{width=\textwidth,keepaspectratio}
\begin{tabular}{c|c|ccccccccccccccc|c}

\toprule
& & \multicolumn{15}{c|}{\textbf{Zero-shot datasets}} & \\
\cline{3-17} 
\textbf{Methods} &\textbf{\makecell[c]{\small\rotatebox{75}{Tiny-ImageNet}}}  & \textbf{\makecell[c]{\small\rotatebox{75}{CIFAR-10}}} & \textbf{\makecell[c]{\small\rotatebox{75}{CIFAR-100}}} & \textbf{\makecell[c]{\small\rotatebox{75}{STL-10}}} & \textbf{\makecell[c]{\small\rotatebox{75}{SUN397}}} & \textbf{\makecell[c]{\small\rotatebox{75}{Food101}}} & \textbf{\makecell[c]{\small\rotatebox{75}{Oxfordpets}}} & \textbf{\makecell[c]{\small\rotatebox{75}{Flowers102}}} & \textbf{\makecell[c]{\small\rotatebox{75}{DTD}}}   & \textbf{\makecell[c]{\small\rotatebox{75}{EuroSAT}}} & \textbf{\makecell[c]{\small\rotatebox{75}{FGVC-Aircraft}}} & \textbf{\makecell[c]{\small\rotatebox{75}{ImageNet}}} & \textbf{\makecell[c]{\small\rotatebox{75}{Caltech-101}}} & \textbf{\makecell[c]{\small\rotatebox{75}{Caltech-256}}} & \textbf{\makecell[c]{\small\rotatebox{75}{StanfordCars}}} & \textbf{\makecell[c]{\small\rotatebox{75}{PCAM}}}  &\(\boldsymbol{A_{\textrm{Robust}}}\) \\

\midrule
CLIP~\cite{CLIP} &0.21	&0.36	&0.10	&10.59	&1.16	&0.82	&1.23	&1.09	&2.18	&0.01	&0.00	&1.14	&13.50	&7.36	&2.36	&0.07	&3.64 \\ 

FT-Clean &1.26 &0.98 &0.14 &9.98 &1.09 &2.05 &0.49 &0.75 &2.50 &0.03 &0.03 &1.22 &7.93 &5.16 &1.42 &1.93 &2.20 \\

FT-Adv. &51.65 &38.49 &20.97 &69.29 &16.87 &11.54 &34.78 &18.82 &15.05 &11.77 &1.74 &16.86 &50.48 &39.78 &8.02 &\textbf{48.78} &28.43\\

TeCoA~\cite{TeCoA} &35.74 &30.14 &16.76 &66.35 &18.20 &13.36 &36.33 &21.17 &15.11 &11.58 &1.50 &16.86 &54.86 &40.62 &8.07 &47.86 &27.16\\

FARE~\cite{FARE} &24.40 &21.35 &9.98 &60.38 &8.55 &10.13 &25.31 &16.08 &12.39 &1.80 &1.09 &9.68 &45.46 &34.07 &4.13 &10.75 &18.47\\

LAAT~\cite{LAAT} &52.88 &35.56 &21.54 &67.56 &18.25 &12.38 &35.73 &18.54 &11.01 &12.74 &1.47 &18.08 &55.78 &41.28 &6.93 &37.46 &27.95\\

PMG-AFT~\cite{PMG-AFT} &44.59	&44.86	&24.15	&74.11	&19.99	&17.33	&39.88	&20.95	&13.51	&12.09	&1.47	&19.51	&60.99	&44.46	&10.57	&\underline{48.59}	&31.07 \\ 

\rowcolor{LightPurple}
 TGA-ZSR (ours) &\makecell[c]{\underline{63.73} \\ \scriptsize{($\pm$ 0.65)}} &\makecell[c]{\underline{60.55} \\ \scriptsize{($\pm$ 0.77)}} &\makecell[c]{\underline{34.66} \\ \scriptsize{($\pm$ 0.43)}} &\makecell[c]{\underline{84.46} \\ \scriptsize{($\pm$ 1.08)}}  &\makecell[c]{\underline{31.34} \\ \scriptsize{($\pm$ 1.48)}} &\makecell[c]{\underline{32.84} \\ \scriptsize{($\pm$ 0.94)}}  &\makecell[c]{\underline{56.97} \\ \scriptsize{($\pm$ 1.98)}} &\makecell[c]{\underline{33.05} \\ \scriptsize{($\pm$ 0.62)}}  &\makecell[c]{\underline{21.13} \\ \scriptsize{($\pm$ 0.67)}}  &\makecell[c]{\underline{13.83} \\ \scriptsize{($\pm$ 0.44)}} &\makecell[c]{\underline{4.56} \\ \scriptsize{($\pm$ 0.03)}}  &\makecell[c]{\underline{27.41} \\ \scriptsize{($\pm$ 1.23)}} &\makecell[c]{\underline{69.80} \\ \scriptsize{($\pm$ 0.95)}}  &\makecell[c]{\underline{59.48} \\ \scriptsize{($\pm$ 0.44)}} &\makecell[c]{\underline{19.50} \\ \scriptsize{($\pm$ 1.00)}}   &\makecell[c]{47.59 \\ \scriptsize{($\pm$ 0.83)}} 
&\makebox[2pt][c]{\makecell[c]{\underline{41.24} \\ \scriptsize{($\pm$ 0.73)}}} \\ 

\rowcolor{LightPurple}
Comp-TGA (ours) &\makecell[c]{\textbf{68.89} \\ \scriptsize{($\pm$ 0.98)}} &\makecell[c]{\textbf{70.36} \\ \scriptsize{($\pm$ 1.68)}} &\makecell[c]{\textbf{42.07} \\ \scriptsize{($\pm$ 1.15)}} &\makecell[c]{\textbf{86.13} \\ \scriptsize{($\pm$ 0.67)}}  &\makecell[c]{\textbf{34.35} \\ \scriptsize{($\pm$ 1.29)}} &\makecell[c]{\textbf{33.90} \\ \scriptsize{($\pm$ 1.10)}}  &\makecell[c]{\textbf{60.42} \\ \scriptsize{($\pm$ 0.90)}} &\makecell[c]{\textbf{35.72} \\ \scriptsize{($\pm$ 1.47)}}  &\makecell[c]{\textbf{23.07} \\ \scriptsize{($\pm$ 0.52)}}  &\makecell[c]{\textbf{16.67} \\ \scriptsize{($\pm$ 0.75)}} &\makecell[c]{\textbf{5.31} \\ \scriptsize{($\pm$ 0.19)}}  &\makecell[c]{\textbf{29.47} \\ \scriptsize{($\pm$ 1.08)}} &\makecell[c]{\textbf{72.11} \\ \scriptsize{($\pm$ 0.62)}}  &\makecell[c]{\textbf{60.70} \\ \scriptsize{($\pm$ 1.17)}} &\makecell[c]{\textbf{23.07} \\ \scriptsize{($\pm$ 1.47)}}   &\makecell[c]{48.52 \\ \scriptsize{($\pm$ 0.22)}} 
&\makebox[2pt][c]{\makecell[c]{\textbf{44.42} \\ \scriptsize{($\pm$ 0.84)}}} \\

\bottomrule
\end{tabular}
\end{adjustbox}
\end{table*}
	
	\subsection{Main Results}
	To validate the effectiveness of our approaches, we conduct comparisons with several state-of-the-art methods such as TeCoA~\cite{TeCoA}, PMG-AFT~\cite{PMG-AFT}, FARE~\cite{FARE}, LAAT~\cite{LAAT}, and TTC~\cite{TTC}. Additionally, we extend the comparison to include CLIP (the original pre-trained CLIP model), FT-Adv. (adversarial fine-tuning using the contrastive loss of the original CLIP) and FT-Clean (fine-tuning on clean examples with the contrastive loss of the original CLIP) for a comprehensive evaluation. We evaluate the robustness of several approaches against PGD attack with a perturbation bound of $\varepsilon = 1/255$.

	\noindent\textbf{Adversarial Zero-shot Robust Accuracy.}
	Table~\ref{tab:main_result_adv} shows that our proposed method, Comp-TGA, achieves a substantial improvement of 39.56\% in \(A_{\textrm{Robust}}\) over the original CLIP model, outperforming its predecessor TGA-ZSR, which achieves 37.19\%. Compared to the current state-of-the-art method, PMG-AFT, Comp-TGA achieves an improvement of 11.95\% in \(A_{\textrm{Robust}}\), while TGA-ZSR also shows a solid 9.58\% improvement. These results highlight the effectiveness of our progressive design: Built upon TGA-ZSR, Comp-TGA further improves its robustness through the integration of complementary text-guided attention mechanisms. In general, our methods are superior to all the other methods on most datasets except for a comparable result on the PCAM dataset. In addition, we obtain the best result on Tiny-ImageNet, which is not a strict zero-shot test. It indicates that our methods are robust to adversarial attacks on both seen and unseen datasets. 
	
	\noindent\textbf{Zero-shot Clean Accuracy.} 
	Table~\ref{tab:main_result_cle} illustrates the model's accuracy for clean examples using different methods. Our improved method, Comp-TGA, achieves a gain of 8.84\% in \(A_{\textrm{Clean}}\) over PMG-AFT and 1.07\% over FT-Clean, while its predecessor, TGA-ZSR, also yields solid improvements of 9.84\% and 2.07\%, respectively. Similar to Table~\ref{tab:main_result_adv}, zero-shot clean accuracy exhibits improvement not only on an individual dataset but across all datasets. However, we observed that our methods are 4.41\% and 3.41\% lower than that of FARE in \(A_{\textrm{Clean}}\), respectively. It is important to note that FARE prioritizes preserving zero-shot clean accuracy. In contrast, our methods maintain competitive clean accuracy while achieving significantly better robustness. Notably, Comp-TGA, despite slightly lower clean accuracy than TGA-ZSR, achieves stronger adversarial robustness, making it the more balanced and practically effective solution.
	
	\subsection{Experiments on More Attack Types}
	\noindent\textbf{Results against AutoAttack.} AutoAttack~\cite{AutoAttack} stands out as a strong attack method for assessing model robustness. We follow TeCoA and PMG-AFT to verify the perturbation bound $\varepsilon$ of 1/255 in the standard version of AutoAttack. The results are summarized in Table~\ref{tab:autoattack}. We can see that the original CLIP model experienced a significant performance decline, decreasing to 0.09\% in \(A_{\textrm{Robust}}\). While all methods exhibit a performance decline, our approaches remain highly competitive, achieving either the best or second-best results across most datasets. Notably, Comp-TGA surpasses TGA-ZSR in terms of average accuracy, further demonstrating its effectiveness against stronger adversarial attacks.

	\noindent\textbf{Results against CW Attack.}\label{CW} 
	CW attack~\cite{cw} is an optimization-based approach designed to generate small perturbations to input data, causing the model to make incorrect predictions while keeping the perturbed input visually similar to the original. We further evaluate the robustness of our approaches against this challenging attack, using a perturbation bound of $\varepsilon = 1/255$. The results, shown in Table~\ref{tab:cw}, demonstrate that Comp-TGA significantly outperforms the state-of-the-art method PMG-AFT, achieving an improvement of 13.35\% in \(A_{\textrm{Robust}}\). Its predecessor, TGA-ZSR, also achieves a solid gain of 10.17\%.
	
	\noindent\textbf{Results against Adaptive Attack.}
	To better assess robustness under worst-case conditions, we follow the simple yet stronger principle adopted in the prior study~\cite{tramer2020adaptive}. Specifically, we assume a standard white-box threat model, where the adversary has full access to the model architecture, parameters, and any necessary additional information. Under this setting, adversarial training is still conducted using PGD-based perturbations, while evaluation is carried out using constructed adaptive attacks tailored to the model structure and optimization objectives for all methods. Detailed descriptions of the adaptive attack design and experimental settings are provided in Appendix B. Table~\ref{tab: adaptive_attack} shows that under adaptive attacks with \(\epsilon = 1/255\), the performance of all methods drops noticeably. Nevertheless, our method consistently outperforms all baselines across almost all zero-shot datasets. While prior approaches and standard adversarial fine-tuning exhibit limited robustness and poor cross-dataset generalization, TGA-ZSR achieves substantial and stable improvements, indicating stronger resistance to worst-case attacks. Comp-TGA further boosts performance and attains the best overall robustness accuracy \(\boldsymbol{A_{\textrm{Robust}}}\), demonstrating that explicitly preserving robust feature and attention structures leads to more reliable zero-shot robustness. These results confirm that the gains are systematic rather than arising from overfitting to specific attack algorithms or datasets.

	\begin{table*}[htbp]
\centering
\caption{Zero-shot robust accuracy on images attacked with \(\varepsilon\) of 1/255 of adaptive attack. We performed several different methods on Tiny-ImageNet and evaluated across 16 datasets. 
The optimal accuracy is highlighted in \textbf{bold}, while the second-best accuracy is \underline{underlined}.
}
\label{tab: adaptive_attack}
\begin{adjustbox}{width=\textwidth,keepaspectratio}
\begin{tabular}{c|c|ccccccccccccccc|c}

\toprule
& & \multicolumn{15}{c|}{\textbf{Zero-shot datasets}} & \\
\cmidrule(lr){3-17}
\textbf{Methods} &\textbf{\makecell[c]{\small\rotatebox{75}{Tiny-ImageNet}}}  & \textbf{\makecell[c]{\small\rotatebox{75}{CIFAR-10}}} & \textbf{\makecell[c]{\small\rotatebox{75}{CIFAR-100}}} & \textbf{\makecell[c]{\small\rotatebox{75}{STL-10}}} & \textbf{\makecell[c]{\small\rotatebox{75}{SUN397}}} & \textbf{\makecell[c]{\small\rotatebox{75}{Food101}}} & \textbf{\makecell[c]{\small\rotatebox{75}{Oxfordpets}}} & \textbf{\makecell[c]{\small\rotatebox{75}{Flowers102}}} & \textbf{\makecell[c]{\small\rotatebox{75}{DTD}}}   & \textbf{\makecell[c]{\small\rotatebox{75}{EuroSAT}}} & \textbf{\makecell[c]{\small\rotatebox{75}{FGVC-Aircraft}}} & \textbf{\makecell[c]{\small\rotatebox{75}{ImageNet}}} & \textbf{\makecell[c]{\small\rotatebox{75}{Caltech-101}}} & \textbf{\makecell[c]{\small\rotatebox{75}{Caltech-256}}} & \textbf{\makecell[c]{\small\rotatebox{75}{StanfordCars}}} & \textbf{\makecell[c]{\small\rotatebox{75}{PCAM}}}  &\(\boldsymbol{A_{\textrm{Robust}}}\) \\

\midrule
FT-Adv. &51.00 &39.90 &21.24 &68.71 &15.95 &10.38 &34.07 &17.81 &13.46 &11.69 &1.26 &16.16 &51.03 &38.35 &6.11 &42.84 &27.50 \\

TeCoA~\cite{TeCoA} &37.57 &30.30 &17.53 &67.19 &19.70 &14.76 &36.44 &22.46 &17.45 &12.14 &1.62 &18.18 &55.86 &41.88 &8.49 &\underline{47.39} &28.06 \\

FARE~\cite{FARE} &22.64 &20.57 &10.63 &58.39 &8.33 &10.75 &24.45 &15.48 &10.80 &0.15 &0.84 &10.52 &45.51 &34.02 &4.32 &3.10 &17.53 \\

LAAT~\cite{LAAT} &49.27 &34.60 &19.74 &66.13 &17.58 &11.38 &35.00 &18.15 &9.63 &8.84 &1.62 &17.06 &55.13 &40.20 &5.48 &38.08 &26.74 \\

PMG-AFT~\cite{PMG-AFT} &44.19 &44.67 &23.96 &74.05 &19.88 &17.25 &39.85 &20.83 &13.35 &12.03 &1.56 &19.42 &60.74 &44.39 &10.12 &30.81 &29.82 \\

\rowcolor{LightPurple}
TGA-ZSR (ours) &\underline{60.53} &\underline{57.46} &\underline{31.87} &\underline{81.46} &\underline{30.46} &\underline{30.73} &\underline{54.62} &\underline{31.79} &\underline{21.60} &\underline{13.76} &\underline{4.26} &\underline{25.95} &\underline{68.28} &\underline{57.37} &\underline{17.26} &45.35 &\underline{39.55} \\

\rowcolor{LightPurple}
Comp-TGA (ours) &\textbf{66.65} &\textbf{66.84} &\textbf{39.70} &\textbf{84.38} &\textbf{32.63} &\textbf{30.88} &\textbf{56.06} &\textbf{33.68} &\textbf{22.71} &\textbf{16.02} &\textbf{4.56} &\textbf{26.90} &\textbf{70.31} &\textbf{58.93} &\textbf{20.07} &\textbf{48.13} &\textbf{42.40} \\ 

\bottomrule
\end{tabular}
\end{adjustbox}
\end{table*}
	\begin{table*}[t]
\centering
\caption{Comparison with test-time defense (TTC). Zero-shot robust and clean accuracy on 16 datasets. We performed several different methods on Tiny-ImageNet, using $\varepsilon$ of 1/255 of PGD to generate adversarial examples for training, and evaluated them across 16 datasets under PGD~\cite{PGD}, CW~\cite{cw} and AutoAttack~\cite{AutoAttack} with the same attack strength. 
}
\label{tab:test_time}  
\begin{adjustbox}{width=\textwidth,keepaspectratio}
\begin{tabular}{c|c|c|ccccccccccccccc|c}

\toprule
& & & \multicolumn{15}{c|}{\textbf{Zero-shot datasets}} & \\
\cline{4-18} 
\textbf{Attack} &\textbf{Methods} &\textbf{\makecell[c]{\small\rotatebox{75}{Tiny-ImageNet}}}  & \textbf{\makecell[c]{\small\rotatebox{75}{CIFAR-10}}} & \textbf{\makecell[c]{\small\rotatebox{75}{CIFAR-100}}} & \textbf{\makecell[c]{\small\rotatebox{75}{STL-10}}} & \textbf{\makecell[c]{\small\rotatebox{75}{SUN397}}} & \textbf{\makecell[c]{\small\rotatebox{75}{Food101}}} & \textbf{\makecell[c]{\small\rotatebox{75}{Oxfordpets}}} & \textbf{\makecell[c]{\small\rotatebox{75}{Flowers102}}} & \textbf{\makecell[c]{\small\rotatebox{75}{DTD}}}   & \textbf{\makecell[c]{\small\rotatebox{75}{EuroSAT}}} & \textbf{\makecell[c]{\small\rotatebox{75}{FGVC-Aircraft}}} & \textbf{\makecell[c]{\small\rotatebox{75}{ImageNet}}} & \textbf{\makecell[c]{\small\rotatebox{75}{Caltech-101}}} & \textbf{\makecell[c]{\small\rotatebox{75}{Caltech-256}}} & \textbf{\makecell[c]{\small\rotatebox{75}{StanfordCars}}} & \textbf{\makecell[c]{\small\rotatebox{75}{PCAM}}}  &\textbf{Average} \\

\midrule
\multirow{3}{*}{PGD} &TTC~\cite{TTC} &24.06 &30.37 &17.24 &77.63 &\textbf{42.25} &\textbf{61.13} &\textbf{60.73} &\textbf{41.75} &\textbf{29.31} &10.58 &\textbf{13.59} &\textbf{38.99} &65.23 &59.73 &\textbf{35.46} &\textbf{50.13} &41.14\\ 

& TGA-ZSR (ours) &\underline{63.95} &\underline{61.45} &\underline{35.27} &\underline{84.22}  &33.22  &\underline{33.97}  &57.75  &34.55  &22.08  &\underline{14.27} &4.75 &28.74  &\underline{70.97}  &\underline{60.06} &20.40 &47.76   &\underline{42.09} \\ 

& Comp-TGA (ours) &\textbf{69.22}  &\textbf{70.46} &\textbf{42.26} &\textbf{86.23}   &\underline{34.61} &33.57 &\underline{60.09} &\underline{35.84}  &\underline{23.90}  &\textbf{16.64}  &\underline{5.41} &\underline{29.47}  &\textbf{72.16}  &\textbf{60.83}  &\underline{22.17}  &\underline{48.56}  &\textbf{44.46} \\ 

\midrule

\multirow{3}{*}{CW} &TTC~\cite{TTC} &21.86 &28.52 &17.22 &73.00 &\textbf{40.13} &\textbf{57.81} &\underline{58.84} &\textbf{39.18} &\textbf{27.29} &10.50 &\textbf{13.71} &\textbf{36.97} &63.91 &57.94 &\textbf{33.06} &\textbf{50.66} &39.41 \\

&TGA-ZSR (ours) &\underline{63.73} &\underline{60.55} &\underline{34.66}  &\underline{84.46}  &31.34 &32.84 &56.97 &33.05  &21.13 &\underline{13.83} &4.56 &27.41  &\underline{69.80} &\underline{59.48} &19.50 &47.59 &\underline{41.24}\\ 

& Comp-TGA (ours) &\textbf{68.89} &\textbf{70.36}  &\textbf{42.07}  &\textbf{86.13}   &\underline{34.35}  &\underline{33.90} &\textbf{60.42}  &\underline{35.72}   &\underline{23.07}   &\textbf{16.67} &\underline{5.31}   &\underline{29.47}  &\textbf{72.11}  &\textbf{60.70} &\underline{23.07}  &\underline{48.52} &\textbf{44.42} \\

\midrule

\multirow{3}{*}{AutoAttack} &TTC~\cite{TTC} &4.93 &3.13 &6.77 &5.51 &7.19 &5.05 &6.41 &10.46 &8.03 &\textbf{15.89} &\textbf{3.51} &6.51 &10.25 &7.26 &4.96 &5.65 &6.97 \\

& TGA-ZSR (ours) &\underline{49.49} &\underline{40.69} &\underline{22.30} &\underline{71.97} &\underline{20.14} &\textbf{15.41} &\underline{39.82} &\underline{21.35} &\underline{16.81} &11.31 &2.34 &\underline{19.22}  &\textbf{57.65} &\textbf{45.73} &\underline{10.14} &\textbf{48.01} &\underline{30.77} \\

& Comp-TGA (ours) &\textbf{53.00} &\textbf{42.18} &\textbf{23.58} &\textbf{73.04}  &\textbf{20.47} &\underline{15.30}  &\textbf{40.28} &\textbf{22.37} &\textbf{17.17}  &\underline{12.28} &\underline{3.43} &\textbf{19.47} &\underline{57.58} &\underline{45.11} &\textbf{10.24}  &\underline{47.05} &\textbf{31.41}\\

\midrule

\multirow{3}{*}{Clean} &TTC~\cite{TTC} &51.53 &\underline{85.39} &\textbf{59.57} &\textbf{96.65} &\textbf{53.51} &\textbf{82.25} &\textbf{83.02} &\textbf{64.33} &\textbf{36.70} &\textbf{53.21} &\textbf{17.94} &\textbf{48.70} &\textbf{85.59} &\textbf{79.58} &\textbf{47.98} &\textbf{57.74} &\textbf{62.73}\\ 

& TGA-ZSR (ours) &\underline{75.72} &\textbf{86.46} &\underline{56.52} &\underline{93.48} &\underline{51.99} &\underline{57.59}  &\underline{77.32} &\underline{48.08} &29.06 &\underline{24.24} &\underline{11.93} &\underline{48.04} &\underline{80.70} &\underline{74.74} &\underline{36.62} &\underline{49.58} &\underline{56.44} \\ 

& Comp-TGA (ours) &\textbf{76.07} &84.46 &55.64 &92.88 &51.47 &54.01 &76.49 &46.07 &\underline{29.43}  &23.25 &11.30 &46.08 &80.50 &73.31 &36.55 &49.46 &55.44\\ 
\midrule

\multirow{3}{*}{\(A_{Overall}\)} &TTC~\cite{TTC} &25.60 &36.85 &25.20 &63.20 &\textbf{35.77} &\textbf{51.56} &52.25 &\textbf{38.93} &\textbf{25.33} &\textbf{22.55} &\textbf{12.19} &\textbf{32.79} &56.24 &51.13 &\textbf{30.36} &41.04 &37.56 \\

& TGA-ZSR (ours) &\underline{63.22} &\underline{62.29} &\underline{37.19} &\underline{83.53} &34.17 &\underline{34.95} &\underline{57.97} &34.26 &22.27 &15.91 &5.90 &30.85 &\underline{69.78} &\textbf{60.00} &21.67 &\underline{48.24} &\underline{42.64} \\

& Comp-TGA (ours) &\textbf{66.80} &\textbf{66.87} &\textbf{40.89} &\textbf{84.57} &\underline{35.23} &34.20 &\textbf{59.32} &\underline{35.00} &\underline{23.39} &\underline{17.21} &\underline{6.36} &\underline{31.12} &\textbf{70.59} &\underline{59.99} &\underline{23.01} &\textbf{48.40} &\textbf{43.93} \\

\bottomrule
\end{tabular}
\end{adjustbox}
\vspace{-2mm}
\end{table*}
	\begin{table*}[t]
% \begin{table}[ht]
\centering
\caption{Comparison of vision-based attention and our text-guided attention. We evaluate the state-of-the-art method PMG-AFT alongside our pipeline, incorporating two different types of attention mechanisms on Tiny-ImageNet and evaluating performance across 16 datasets.}
\label{tab:atten_adv}  
\begin{adjustbox}{width=\textwidth,keepaspectratio}
\begin{tabular}{c|c|c|ccccccccccccccc|c}

\toprule
& & & \multicolumn{15}{c|}{\textbf{Zero-shot datasets}} & \\
\cline{4-18} 
\textbf{Test} &\textbf{Methods} &\textbf{\makecell[c]{\small\rotatebox{75}{Tiny-ImageNet}}}  & \textbf{\makecell[c]{\small\rotatebox{75}{CIFAR-10}}} & \textbf{\makecell[c]{\small\rotatebox{75}{CIFAR-100}}} & \textbf{\makecell[c]{\small\rotatebox{75}{STL-10}}} & \textbf{\makecell[c]{\small\rotatebox{75}{SUN397}}} & \textbf{\makecell[c]{\small\rotatebox{75}{Food101}}} & \textbf{\makecell[c]{\small\rotatebox{75}{Oxfordpets}}} & \textbf{\makecell[c]{\small\rotatebox{75}{Flowers102}}} & \textbf{\makecell[c]{\small\rotatebox{75}{DTD}}}   & \textbf{\makecell[c]{\small\rotatebox{75}{EuroSAT}}} & \textbf{\makecell[c]{\small\rotatebox{75}{FGVC-Aircraft}}} & \textbf{\makecell[c]{\small\rotatebox{75}{ImageNet}}} & \textbf{\makecell[c]{\small\rotatebox{75}{Caltech-101}}} & \textbf{\makecell[c]{\small\rotatebox{75}{Caltech-256}}} & \textbf{\makecell[c]{\small\rotatebox{75}{StanfordCars}}} & \textbf{\makecell[c]{\small\rotatebox{75}{PCAM}}}  &\textbf{Average} \\

\midrule
 \multirow{4}{*}{Robust} &PMG-AFT\cite{PMG-AFT} &47.11 &46.01 &25.83 &74.51 &22.21 &19.58 &41.62 &23.45 &15.05 &12.54 &1.98 &21.43 &62.42 &45.99 &11.72 &\textbf{48.64} &32.51 \\ 

&Vision-based &52.81 &40.46 &22.66 &70.26 &19.50 &13.74 &37.67 &19.78 &16.97 &11.79 &2.64 &18.08 &55.64 &42.45 &8.88 &38.11 &29.47 \\ 

&\cellcolor{LightPurple}TGA-ZSR (ours) &\cellcolor{LightPurple}63.95 &\cellcolor{LightPurple}61.45 &\cellcolor{LightPurple}35.27 &\cellcolor{LightPurple}84.22 &\cellcolor{LightPurple}33.22 &\cellcolor{LightPurple}\textbf{33.97} &\cellcolor{LightPurple}57.75 &\cellcolor{LightPurple}34.55 &\cellcolor{LightPurple}22.08 &\cellcolor{LightPurple}14.27 &\cellcolor{LightPurple}4.75 &\cellcolor{LightPurple}28.74 &\cellcolor{LightPurple}70.97 &\cellcolor{LightPurple}60.06 &\cellcolor{LightPurple}20.40 &\cellcolor{LightPurple}47.76 &\cellcolor{LightPurple}42.09 \\ 

&\cellcolor{LightPurple}Comp-TGA (ours) &\cellcolor{LightPurple}\textbf{69.22} &\cellcolor{LightPurple}\textbf{70.46} &\cellcolor{LightPurple}\textbf{42.26} &\cellcolor{LightPurple}\textbf{86.23} &\cellcolor{LightPurple}\textbf{34.61} &\cellcolor{LightPurple}33.57 &\cellcolor{LightPurple}\textbf{60.09} &\cellcolor{LightPurple}\textbf{35.84} &\cellcolor{LightPurple}\textbf{23.90} &\cellcolor{LightPurple}\textbf{16.64} &\cellcolor{LightPurple}\textbf{5.41}  &\cellcolor{LightPurple}\textbf{29.47} &\cellcolor{LightPurple}\textbf{72.16} &\cellcolor{LightPurple}\textbf{60.83} &\cellcolor{LightPurple}\textbf{22.17}  &\cellcolor{LightPurple}48.56 &\cellcolor{LightPurple}\textbf{44.46} \\ 

\midrule
\multirow{4}{*}{Clean}&PMG-AFT\cite{PMG-AFT} &67.11 &74.62 &44.68 &88.85 &37.42 &37.47 &66.34 &35.66 &21.17 &17.76 &4.71 &35.93 &76.70 &61.96 &25.21 &49.99 &46.60 \\ 

&Vision-based &74.31 &70.77 &41.03 &87.24 &36.91 &30.07 &62.52 &33.89 &24.10 &16.26 &5.70 &33.59 &72.35 &59.75 &20.50 &\textbf{51.29} &45.02 \\ 

&\cellcolor{LightPurple}TGA-ZSR (ours) &\cellcolor{LightPurple}75.72 &\cellcolor{LightPurple}\textbf{86.46} &\cellcolor{LightPurple}\textbf{56.52} &\cellcolor{LightPurple}\textbf{93.48} &\cellcolor{LightPurple}\textbf{51.99} &\cellcolor{LightPurple}\textbf{57.59} &\cellcolor{LightPurple}\textbf{77.32} &\cellcolor{LightPurple}\textbf{48.08} &\cellcolor{LightPurple}29.06 &\cellcolor{LightPurple}\textbf{24.24} &\cellcolor{LightPurple}\textbf{11.93} &\cellcolor{LightPurple}\textbf{48.04} &\cellcolor{LightPurple}\textbf{80.70} &\cellcolor{LightPurple}\textbf{74.74} &\cellcolor{LightPurple}\textbf{36.62} &\cellcolor{LightPurple}49.58 &\cellcolor{LightPurple}\textbf{56.44} \\ 

&\cellcolor{LightPurple}Comp-TGA (ours) &\cellcolor{LightPurple}\textbf{76.07} &\cellcolor{LightPurple}84.46 &\cellcolor{LightPurple}55.64 &\cellcolor{LightPurple}92.88 &\cellcolor{LightPurple}51.47 &\cellcolor{LightPurple}54.01  &\cellcolor{LightPurple}76.49 &\cellcolor{LightPurple}46.07 &\cellcolor{LightPurple}\textbf{29.43} &\cellcolor{LightPurple}23.25 &\cellcolor{LightPurple}11.30 &\cellcolor{LightPurple}46.08 &\cellcolor{LightPurple}80.50 &\cellcolor{LightPurple}73.31 &\cellcolor{LightPurple}36.55 &\cellcolor{LightPurple}49.46 &\cellcolor{LightPurple}55.44 \\ 

\bottomrule
\end{tabular}
\end{adjustbox}
\vspace{-2mm}
\end{table*}

	\begin{table}[t]
		\centering
		\caption{Comparison of different prompt-guided attention in Comp-TGA. We trained several different methods on Tiny-ImageNet using PGD with $\varepsilon = 1/255$, and evaluated them across 16 datasets under PGD with the same attack strength.}
		\label{tab:prompt-guided}
		\scriptsize
		\begin{tabular}{p{0.9cm}<{\centering}p{1.1cm}<{\centering}p{0.9cm}<{\centering}p{1.cm}<{\centering}p{1.cm}<{\centering}p{1.cm}<{\centering}}
			\toprule
			\textbf{\makecell[c]{Class\\ Prompts}} & \textbf{\makecell[c]{Non-class \\ Prompts}} & \textbf{\makecell[c]{Other \\ Prompts}} & \(\boldsymbol{A_{\textrm{Robust}}}\) &\(\boldsymbol{A_{\textrm{Clean}}}\) &\(\boldsymbol{A_{\textrm{Overall}}}\)  \\
			\midrule
			\checkmark & & & 41.45 & \textbf{56.48} & 48.98 \\
			\checkmark & &  \checkmark & 43.24 & 55.10 & 49.17 \\
			\rowcolor{LightPurple}
			\checkmark & \checkmark & & \textbf{44.46} & 55.44 & \textbf{49.95} \\
			\bottomrule
		\end{tabular}
	\end{table}
	
	\begin{table*}[htbp]
		\centering
		\caption{Zero-shot robust accuracy under PGD~\cite{PGD} attacks with \(\varepsilon\) = 1/255, 2/255, 4/255, and 8/255 and clean accuracy. We performed several different methods on Tiny-ImageNet, using $\varepsilon$ of 1/255 of PGD to generate adversarial examples for training, and evaluated them across 16 datasets.}
		\label{tab:strength_v2}
		\begin{adjustbox}{width=\textwidth,keepaspectratio}
			\begin{tabular}{c|c|ccccccc >{\columncolor{LightPurple}}c >{\columncolor{LightPurple}}c }
				\toprule
				Test &\(\varepsilon\) &CLIP~\cite{CLIP} &FT-Clean &FT-Adv. &TeCoA~\cite{TeCoA} &FARE~\cite{FARE} &LAAT~\cite{LAAT} &PMG-AFT~\cite{PMG-AFT} &TGA-ZSR (ours) &Comp-TGA (ours) \\
				
				\midrule
				\multirow{4}{*}{Robust} &\(1/255\) &4.90 &7.05 &28.83 &28.06 &18.25 &28.02 &32.51 &\underline{42.09} &\textbf{44.46} \\
				&\(2/255\) &3.25 &5.93 &16.65 &14.20 &4.43 &\underline{17.24} &\textbf{19.64} &11.48 &15.20 \\
				&\(4/255\) &2.81 &5.66 &5.72 &3.97 &2.94 &\textbf{10.27} &5.65 &7.49 &\underline{9.96} \\
				&\(8/255\) &2.42 &5.42 &2.20 &1.66 &2.64 &\underline{8.49} &0.82 &5.95 &\textbf{8.61} \\
				\midrule
				Clean &— &\textbf{64.42} & 54.37 &43.36 &45.81 &\underline{59.85} &43.10 &46.60 &56.44 &55.44 \\
				\midrule
				Average &— &15.56 &15.69 &19.35 &18.74 &17.62 &21.42 &21.04 &\underline{24.69} &\textbf{26.73} \\
				
				\bottomrule
			\end{tabular}
		\end{adjustbox}
	\end{table*}
	
	\begin{table}[t]
		\centering
		\caption{Ablation study of each component in Comp-TGA. We trained several different methods on Tiny-ImageNet using PGD with $\varepsilon = 1/255$, and evaluated them across 16 datasets under PGD with the same attack strength.}
		\label{tab:component}
		\small
		\begin{tabular}{p{2cm}p{1.6cm}<{\centering}p{1.6cm}<{\centering}p{1.6cm}<{\centering}}
			\toprule
			\textbf{Methods} &\(\boldsymbol{A_{\textrm{Robust}}}\) &\(\boldsymbol{A_{\textrm{Clean}}}\) &\(\boldsymbol{A_{\textrm{Overall}}}\) \\
			\midrule
			CLIP & 4.90 & 64.42 & 34.66 \\
			\midrule
			+ $L_{\mathrm{CE}}$ & 29.45 & 44.97 & 37.21 \\
			\midrule
			\textbf{Comp-TGA} & & & \\
			+ $L_{\textrm{LARM}}$ & 30.69 & 46.68 & 38.69 \\
			+ $L_{\textrm{GACM}}$ & 44.46 & 55.44 & 49.95 \\
			\bottomrule
		\end{tabular}
	\end{table}

	\subsection{Comparison with Test-Time Defense}
	Our proposed methods are based on fine-tuning, and thus, the previous experiments primarily focused on comparisons with similar methods, such as TeCoA~\cite{TeCoA}, FARE~\cite{FARE}, LAAT~\cite{LAAT}, and PMG-AFT~\cite{PMG-AFT}. However, to provide a more comprehensive evaluation of our approaches, we also compare them with the TTC defense method, which does not require training, using a perturbation bound of 1/255. As shown in the experiments in Table~\ref{tab:test_time}, due to the fact that TTC does not involve training or model modification, it achieves relatively high clean accuracy\textemdash only 1.69\% lower in \(A_{\textrm{Clean}}\) compared to the original CLIP, and 6.29\% and 7.29\% higher than our methods, TGA-ZSR and Comp-TGA, respectively. However, the lack of training also results in poor robustness under adversarial attacks. In particular, under the AutoAttack setting, TTC performs 24.44\% and 23.80\% worse in \(A_{\textrm{Robust}}\) compared to our methods. More importantly, when considering the overall robustness across all three attack types along with clean accuracy, our method Comp-TGA achieves the best trade-off, outperforming TTC by 6.37\% in \(A_{\textrm{Overall}}\), while TGA-ZSR also outperforms TTC by 5.08\%.

	\subsection{Ablation Study}
	\noindent\textbf{Different Types of Attentions.}
	To comprehensively evaluate the contributions of various attention mechanisms, we conducted experiments comparing vision-based attention and different types of prompt-guided attention under PGD attack with a perturbation bound of 1/255.
	\subsubsection{Comparison of vision-based attention and text-guided attention} To validate the important role of text-guided attention in our methods, we conducted experiments by replacing it with vision-based attention. We employ Grad-CAM~\cite{Grad-cam}, a widely adopted method, for generating attention maps based on vision. Table~\ref{tab:atten_adv} demonstrates that replacing the text-guided attention with vision-based attention yields results that are still comparable to the state-of-the-art method PMG-AFT in terms of both zero-shot robust accuracy and clean accuracy. This finding validates the effectiveness of the overall pipeline of our methods. Furthermore, our text-guided attention significantly improves the average accuracy, demonstrating the advantage of incorporating textual guidance. 
	
	\subsubsection{Comparison of different prompt-guided attention in Comp-TGA}
	Comp-TGA integrates two types of foreground attention by two distinct prompts. Therefore, to evaluate the contributions of different prompt-guided attention, we conducted experiments using Class Prompts, Non-Class Prompts, and Other Prompts (e.g. ``This is the background of the photo''). These experiments presented in Table~\ref{tab:prompt-guided} demonstrate the advantages of complementary attention fusion through comparisons with the first row and the other two rows. Furthermore, the comparison between the second and third rows highlights the importance of class prompts with non-class prompts.

	\noindent\textbf{Effect of Attack Strength.}
	Table~\ref{tab:strength_v2} provides a comprehensive view of model performance across multiple attack strengths (\(\varepsilon\) = 1/255, 2/255, 4/255, and 8/255). The results show that all methods exhibit decreasing accuracy as \(\epsilon\) increases. LAAT~\cite{LAAT} enhances the differences between category text embeddings through an expansion algorithm, while PMG-AFT~\cite{PMG-AFT} relies on a pre-trained model for guided adversarial fine-tuning. Although both methods achieve some improvements on certain tasks, they incur high computational costs (PMG-AFT memory usage is 18449 Mb compared to our 16265 Mb and 16293 Mb, as reported in the Table~\ref{tab:computattional_overhead}) or long training times (LAAT training time per batch is 3.47 seconds compared to our 0.96 and 1.10 seconds, as reported in the Table~\ref{tab:computattional_overhead}), and they fail to maintain zero-shot generalization and robustness across different attacks and tasks. Their zero-shot generalization (clean accuracy) is typically 8 to 10 points lower than ours, and they struggle to maintain robustness under stronger attacks such as CW, AutoAttack, and APGD, as well as cross-task benchmarks like object detection.\footnote{Additional experiments on APGD are provided in Appendix D, and experiments with object detection in Appendix C.} In general, our methods achieve the highest average robust accuracy with lower computational cost and demonstrate stronger cross-scenario adaptability and scalability.

	\noindent\textbf{Effect of Each Component.}
	We conducted several experiments on Comp-TGA to thoroughly evaluate the effectiveness of each component under PGD attack with a perturbation bound of 1/255, as summarized in Table~\ref{tab:component}. Starting from the CLIP baseline, the low robust accuracy indicates high vulnerability to adversarial perturbations. Using $L_{\textrm{CE}}$ alone significantly enhances the model's robustness through standard adversarial training, but it also results in a notable decrease in clean accuracy compared to the original CLIP model. Comp-TGA incorporates a complementary attention mechanism that fuses multiple text-guided attention sources. This extension consistently yields additional gains. Specifically, integrating the Local Attention Refinement module in Comp-TGA further enhances both metrics by refining attention regions and guiding the model toward semantically relevant areas, which leads to improvements of 1.24\% in \(A_{\textrm{Robust}}\) and 1.71\% in \(A_{\textrm{Clean}}\). When the Global Attention Constraint module is further added, Comp-TGA achieves an additional 13.77\% gain in \(A_{\textrm{Robust}}\) and 10.47\% in \(A_{\textrm{Clean}}\), demonstrating that enforcing attention consistency between clean and adversarial examples stabilizes model behavior and achieves a better robustness-generalization balance.

\begin{figure}[t]
		\centering
		\includegraphics[width=0.49\textwidth]{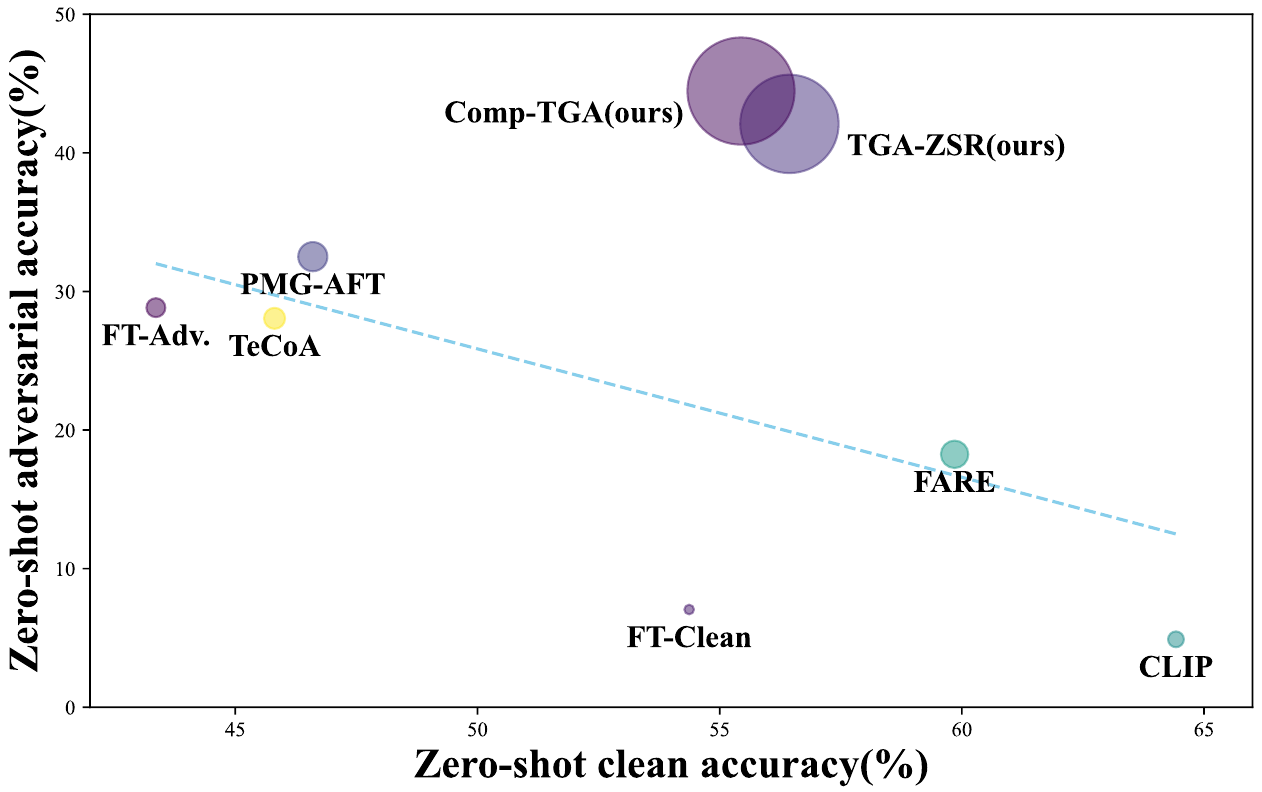}
		\caption{The trade-off between robustness and clean accuracy. Each point on the graph represents a method, with the size of the point quantitatively reflecting the trade-off between robustness and clean accuracy. Larger points indicate methods that achieve a more favorable balance between robustness and clean accuracy.}
		\label{fig:trade-off}
	\end{figure}

	\noindent\textbf{Trade-off between Robust and Clean Accuracy.}
	Achieving the balance between robustness and clean accuracy is crucial in adversarial training. Overfitting in models tends to yield high robustness but low clean accuracy, whereas underfitting typically results in the opposite scenario. As shown in Fig.~\ref{fig:trade-off} \footnote{The point size is computed as \(size = s \times (A_{\textrm{Robust}} + A_{\textrm{Clean}})\), where \(s\) is a scaling factor applied consistently to all methods to ensure that the sizes are comparable across the figure.}, methods positioned close to the dotted line excel in either adversarial accuracy or clean accuracy, yet they often struggle to strike a balance between robustness and clean accuracy. In contrast, our methods demonstrate not only an enhancement in the model's robustness but also the maintenance of clean accuracy, resulting in an overall superior performance.
	
	\subsection{Computational Overhead and Time Efficiency}\label{computational}
	We have evaluated our methods against others in terms of memory usage, training time, and test time, and the results are summarized in Table~\ref{tab:computattional_overhead}. Among the baselines, FARE~\cite{FARE} stands out for its efficiency, largely because it adopts a lightweight fine-tuning strategy that avoids text processing and dual-branch structures. In contrast, TGA-ZSR and Comp-TGA incorporate text-guided attention and a dual-branch design to improve robustness and preserve generalization. This additional structure naturally increases computational cost and leads to longer training times than FARE. However, our methods achieve substantially stronger robustness while maintaining clean accuracy. The performance gap becomes more pronounced under stronger attacks, where FARE degrades more severely. This indicates that the additional training time in our methods is not redundant but supports a significantly better robustness generalization balance. Comp-TGA further strengthens this balance, improving performance over TGA-ZSR while keeping memory usage nearly identical and increasing training time only slightly. At test time, all methods have comparable inference cost except for TTC~\cite{TTC}. Although TTC requires no training, it generates perturbations online during inference, which results in a substantially longer test-time latency than the other approaches.
	
	It is worth noting that to improve the scalability and memory efficiency of our methods, we optimized the original implementation with a particular focus on reducing memory consumption and improving computational efficiency. These improvements do not alter the core methodology or theoretical foundations of the approach, but rather enhance its implementation efficiency. Compared to the conference version of TGA-ZSR~\cite{TGA-ZSR}, the optimized implementation of TGA-ZSR reduces memory usage during training by 23.37\% and improves training efficiency by 14.94\%.

	\begin{table}[t]
		\centering
		\caption{Comparison of memory usage, training time, and overall accuracy.}
		{\small
			\label{tab:computattional_overhead}  
				\begin{tabular}{p{2.4cm}<{\centering}|p{1.4cm}<{\centering}p{1.6cm}<{\centering}p{1.6cm}<{\centering}}
					\toprule
					\multirow{2}{*}{\textbf{Methods}} & \textbf{Memory} & \textbf{Train time} & \textbf{Overall} \\
					& \textbf{usage} & \textbf{(per batch)} & \textbf{Accuracy} \\
					\midrule
					CLIP~\cite{CLIP}    &0 Mb        &0.00s     &34.66\% \\
					FT-Clean            &6561 Mb     &0.21s     &30.71\% \\
					FT-Adv.             &10859 Mb    &0.60s     &36.10\% \\
					TeCoA~\cite{TeCoA}  &12873 Mb    &0.65s     &36.94\% \\
					FARE~\cite{FARE}    &9821 Mb      &0.25s    &39.05\% \\
					LAAT~\cite{LAAT}    &15527 Mb    &3.47s     &35.56\% \\
					PMG-AFT~\cite{PMG-AFT} & 18449 Mb &1.06s    &39.56\% \\
					\rowcolor{LightPurple}
					TGA-ZSR (ours)      &16265 Mb   &0.96s     &49.27\% \\ 
					\rowcolor{LightPurple}
					Comp-TGA (ours)    &16293 Mb   &1.10s     &49.95\% \\ 
					\bottomrule
			\end{tabular}}
	\end{table}

	\section{Conclusion and Limitations}\label{discussion}
	In this paper, we discovered that adversarial attacks lead to a shift in text-guided attention.  Building on this observation, we introduce a text-guided approach, TGA-ZSR, which incorporates two key components to perform adversarial fine-tuning and constrain the model. This strategy prevents model drift while enhancing model robustness. Extensive experiments validate the performance of TGA-ZSR, which not only improves CLIP's zero-shot adversarial robustness but also maintains zero-shot clean accuracy on clean examples, gaining a favorable balance. However, we observe that text-guided attention occasionally attends to irrelevant features, resulting in suboptimal performance. Therefore, we further propose a novel approach called Comp-TGA, which integrates two complementary types of foreground attention. This allows the model to capture a more accurate representation of the foreground.
	
	\noindent\textbf{Limitations.} We use a simple text-guided attention mechanism by multiplying the text embedding and vision embedding, which is effective against most attack types. However, for more challenging attacks such as AutoAttack, the improvement remains limited. Moreover, while we provide initial evaluations with adaptive attacks, which are crucial for faithfully assessing worst-case robustness, these attacks may require further refinement to fully capture the model's true vulnerability. This indicates that while our approaches show promise, they may require further refinement to enhance robustness under stronger adversarial scenarios.
	
	\noindent\textbf{Border Impact.} 
	Large-scale pre-trained vision-language models (VLMs) like CLIP~\cite{CLIP} integrate visual and textual data, revolutionizing applications such as image classification, semantic segmentation, and vision question answering. While these models excel in zero-shot learning and transfer learning, they are vulnerable to adversarial attacks, posing risks in critical applications like autonomous vehicles and medical diagnosis. Adversarial training improves robustness but has practical challenges, including increased computational overhead and potential overfitting. Exploring zero-shot adversarial robustness is essential to ensure reliability. 
	
	\bibliography{reference}
	\bibliographystyle{IEEEtran}
	
	\begin{IEEEbiography}[{\includegraphics[width=1in,height=1.25in,clip,keepaspectratio]{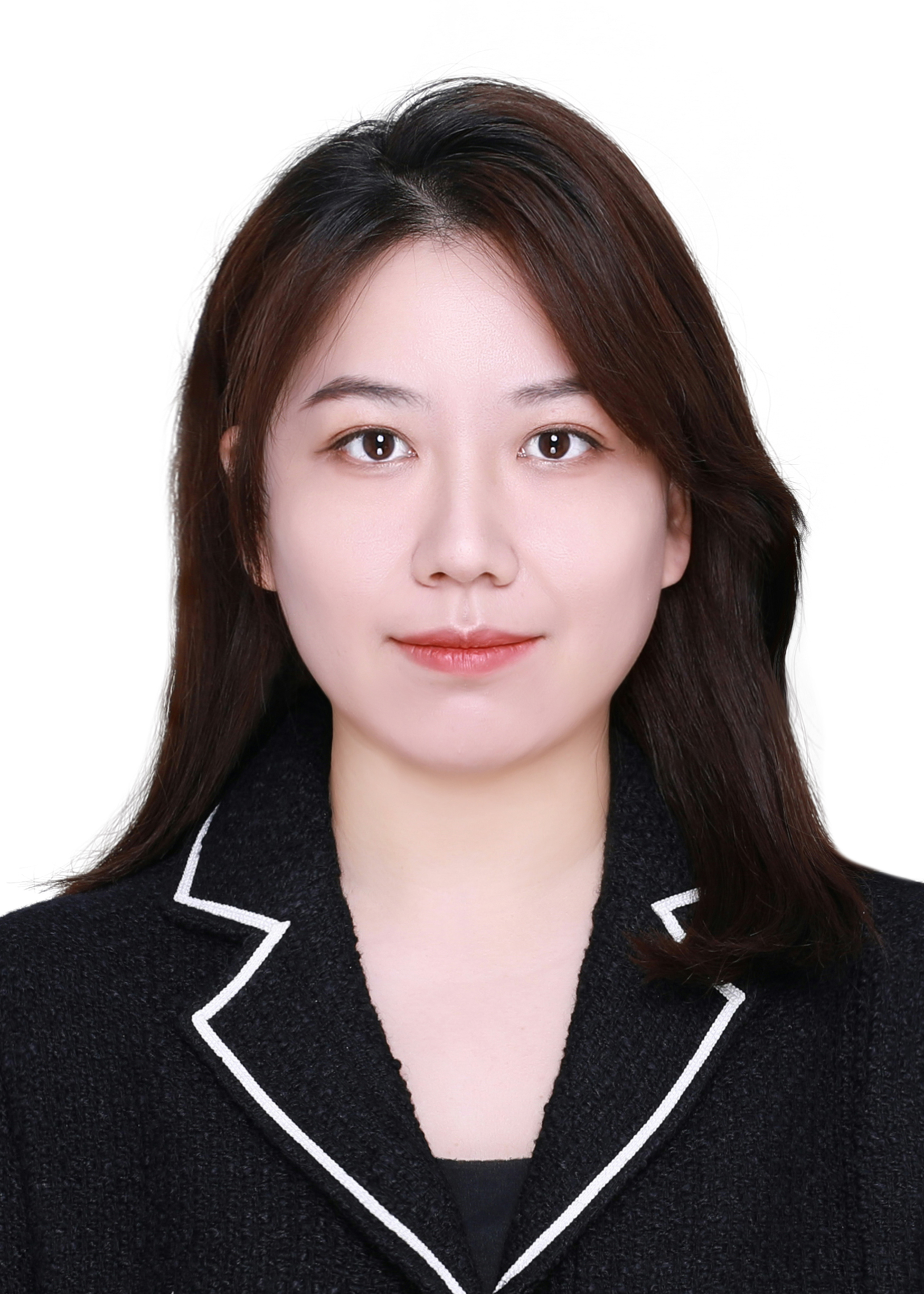}}]{Lu Yu} is currently an associate professor at Tianjin University of Technology, Tianjin, China. Before that She was a post-doc at Heriot Watt University, Edinburgh, UK. She received her Ph.D in computer science from Autonomous University of Barcelona, Barcelona, Spain in 2019 and master degree from Northwestern Polytechnical University in 2015, Xi'an, China.  Her research interests include continual learning, metric learning, multi-model learning and color representation learning.
	\end{IEEEbiography}
	
	\begin{IEEEbiography}[{\includegraphics[width=1in,height=1.25in,clip,keepaspectratio]{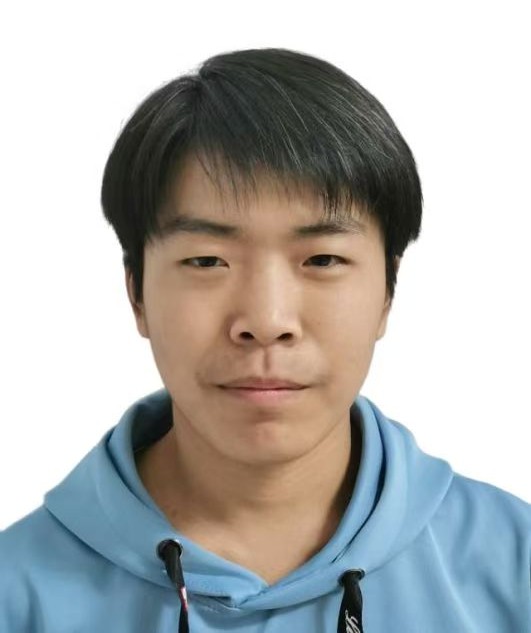}}]{Haiyang Zhang} received the bachelor’s degree in Mathematics and Applied Mathematics from Tianjin University of Commerce, Tianjin, China, in 2022. Currently, he is pursuing the Ph.D. degree at Tianjin University of Technology, Tianjin, China. His research interests include deep learning and adversarial robustness.
	\end{IEEEbiography}

	\begin{IEEEbiography}[{\includegraphics[width=1in,height=1.25in,clip,keepaspectratio]{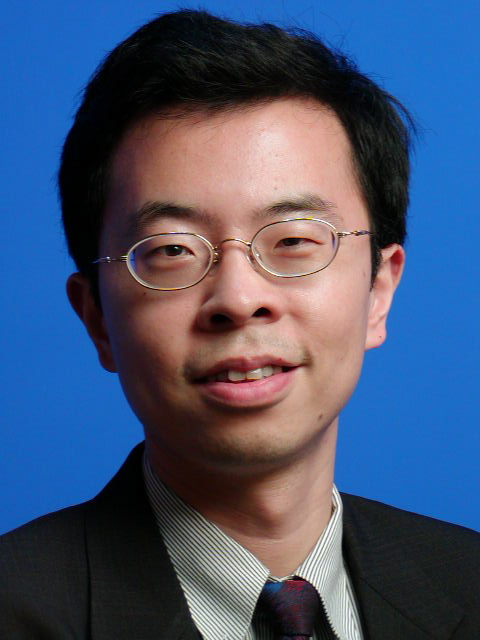}}]{Changsheng Xu} (M’97–SM’99–F’14) is a Professor in State Key Laboratory of Multimodal Artificial Intelligence Systems (MAIS), Institute of Automation, Chinese Academy of Sciences. His research interests include multimedia content analysis/indexing/retrieval, pattern recognition and computer vision. He has held 50 granted/pending patents and published over 400 refereed research papers in these areas. Dr. Xu has served as associate editor, guest editor, general chair, program chair, area/track chair and TPC member for over 20 IEEE and ACM prestigious multimedia journals, conferences and workshops, including IEEE Trans. on Multimedia, ACM Trans. on Multimedia Computing, Communications and Applications and ACM Multimedia conference. He is IEEE Fellow, IAPR Fellow and ACM Distinguished Scientist.
	\end{IEEEbiography}

	\appendices
	\newpage
	
\setcounter{section}{0}
\setcounter{figure}{0}
\setcounter{table}{0}
\setcounter{page}{1}
\setcounter{equation}{0}
\renewcommand{\thetable}{A\arabic{table}}
\renewcommand{\thefigure}{A\arabic{figure}}

\section{Quantitative mIoU of different types of attention}
To quantitatively evaluate the quality of the attention maps, we intend to \textbf{measure their alignment with ground-truth regions using the Intersection over Union (IoU) metric}. However, our experiments primarily focus on image classification tasks, which do not inherently provide segmentation labels. Therefore, standard metrics from semantic segmentation cannot be directly used to evaluate the quality of attention. To demonstrate the advantages of text-guided attention and complementary text-guided attention, we establish standardized statistics and provide quantitative metrics on \textbf{the validation sets of ImageNet-S-50, ImageNet-S-300, and ImageNet-S-919}~\cite{gao2022large}, which include ground truth masks for semantic segmentation tasks. This allows us to quantitatively analyze the spatial focus of the attention maps.

In conventional semantic segmentation tasks, the IoU is typically used to evaluate the accuracy of predicted masks. For a binary predicted mask \(\hat{\mathbf{M}} \in \{0,1\}^p\) and the ground truth mask \(\mathbf{M} \in \{0,1\}^p\), IoU is defined as:
\begin{equation}
	\mathrm{IoU} = \frac{|\hat{\mathbf{M}} \cap \mathbf{M}|}{|\hat{\mathbf{M}} \cup \mathbf{M}|} = \frac{\sum_{i,j}\hat{\mathbf{M}}_{ij}\mathbf{M}_{ij}}{\sum_{ij}\hat{\mathbf{M}}_{ij} + \mathbf{M}_{ij} - \hat{\mathbf{M}}_{ij}\mathbf{M}_{ij}}
\end{equation}

However, our attention maps \(\mathbf{A}\) are continuous values in \([0,1]\), representing the model's attention strength at each spatial location rather than binary predictions. This makes the standard IoU inapplicable and could result in substantial information loss. To quantify how well the attention focuses on the target regions while preserving continuous information, we adopt a soft Jaccard IoU~\cite{wang2023jaccard} metric:
\begin{equation}
	\mathrm{softIoU} = \frac{\sum_{i,j}\mathbf{A}_{ij}\mathbf{M}_{ij}}{\sum_{ij}\mathbf{A}_{ij} + \mathbf{M}_{ij} - \mathbf{A}_{ij}\mathbf{M}_{ij}}
\end{equation}
This metric compares the weighted intersection and union between the attention map and the ground truth mask, providing a smooth measure of attention distribution over the target region and avoiding errors introduced by simple binarization. Moreover, the soft IoU can be averaged over the entire validation set to obtain an overall evaluation of the model's spatial attention distribution.

The experimental results are shown in Table~\ref{tab:iou}. We evaluate different types of attention, namely text-guided attention (TGA) and complementary text-guided attention (Comp-TGA), on both the original CLIP model (ori.) and the finetuned CLIP model (tar.). Overall, the IoU in the finetuned CLIP model is higher than in the original model. For TGA, the IoU on adversarial examples is lower than on clean examples, supporting our claim that adversarial attacks can shift text-guided attention, as illustrated in Fig. 1. Aligning attention based on this observation improves model performance, reflected by the higher IoU in the finetuned CLIP model compared to the original model, demonstrating the effectiveness of our approach. Our improved method, Comp-TGA, further increases the IoU relative to TGA, enabling the model to focus more accurately on the correct object, consistent with Fig. 4.
\begin{table*}[htbp]
	\centering
	\caption{Quantitative mIoU of different types of attention on ImageNet-S-50, ImageNet-S-300, and ImageNet-S-919~\cite{gao2022large}.}
	\label{tab:iou}
	\begin{adjustbox}{width=\textwidth,keepaspectratio}
		\begin{tabular}{c| cccc| cccc| cccc}
			\toprule
			\multicolumn{1}{c|}{\multirow{3}{*}{Type}}  & \multicolumn{4}{c|}{ImageNet-S-50} & \multicolumn{4}{c|}{ImageNet-S-300} &\multicolumn{4}{c}{ImageNet-S-919}\\
			\cmidrule(lr){2-5} \cmidrule(lr){6-9} \cmidrule(lr){10-13}
			&\multicolumn{2}{c}{TGA} &\multicolumn{2}{c|}{Comp-TGA} &\multicolumn{2}{c}{TGA} &\multicolumn{2}{c|}{Comp-TGA} &\multicolumn{2}{c}{TGA} &\multicolumn{2}{c}{Comp-TGA} \\
			\cmidrule(lr){2-3} \cmidrule(lr){4-5} \cmidrule(lr){6-7} \cmidrule(lr){8-9} \cmidrule(lr){10-11} \cmidrule(lr){12-13}
			&ori. &tar. &ori. &tar. &ori. &tar. &ori. &tar. &ori. &tar. &ori. &tar. \\
			\midrule
			Clean Examples &0.3104 &0.3117 &0.3630 &0.3648 &0.3090 &0.3078 &0.3662 &0.3677 &0.2907 &0.2914 &0.3509 &0.3568 \\
			Adversarial Examples &0.3033 &0.3121 &0.3632 &0.3646 &0.2980 &0.3064 &0.3657 &0.3676 &0.2834 &0.2903 &0.3516 &0.3569 \\
			\bottomrule
		\end{tabular}
	\end{adjustbox}
\end{table*}

\section{Experiments on The Adaptive Attack}
To better assess robustness under worst-case conditions, we therefore revisit our evaluation protocol, following the simple yet stronger principle adopted in the prior study~\cite{tramer2020adaptive}. Specifically, we assume a standard white-box threat model, where the adversary has full access to the model architecture, parameters, and any necessary additional information. Under this setting, adversarial training is still conducted using PGD-based perturbations, while evaluation is carried out using constructed adaptive attacks tailored to the model structure and optimization objectives for all methods. These adaptive attacks directly target the actual training objectives and inference pathways of the models, thereby reducing the potential biases induced by relying on a single gradient-based attack algorithm. The detailed design of the adaptive attacks and analysis are presented below.

To align with the adaptive attack principle~\cite{tramer2020adaptive}, we design attack objectives that are explicitly matched to the defense strategies of different methods, rather than relying on a single fixed loss. Based on the form of the optimization objective, the adaptive attacks used in our evaluation can be broadly categorized into two types: Cross-Entropy-based attacks and Log-Sum-Exp-based attacks.

\textbf{Cross-Entropy-based} attacks directly optimize the cross-entropy loss and are naturally aligned with defenses whose training objectives are explicitly classification-driven or rely on prompt-level modifications. In such cases, cross-entropy provides a faithful and sufficiently strong adaptive objective that directly targets the model's decision boundary.

\noindent $\bullet$ \textbf{TeCoA}~\cite{TeCoA} fine-tunes CLIP on adversarial examples using the cross-entropy loss. During evaluation, adversarial examples are generated by directly optimizing the cross-entropy loss, which corresponds to the standard PGD~\cite{PGD} method.
	\begin{equation}
		\max_{\|\delta\| \leq \epsilon} L_{\mathrm{CE}}(\textbf{x} + \delta, \textbf{t}, \textbf{y})
	\end{equation}
	where \(\delta\) denotes the adversarial perturbation bounded by \(\epsilon\), \(\textbf{x} + \delta\) denotes the adversarial example, \(\textbf{t}\) denotes the text prompt, \(\textbf{y}\) is the ground-truth label and \(L_{\mathrm{CE}}\) is the cross-entropy loss.

\noindent $\bullet$ \textbf{LAAT}~\cite{LAAT} enhances adversarial robustness through a textual expansion algorithm that augments the original text prompts. To properly evaluate its robustness, we design a prompt-aware adaptive attack that operates from the text prompt perspective. During evaluation, adversarial examples are generated by directly optimizing the cross-entropy loss with respect to the modified text prompts, thereby explicitly targeting the textual modification mechanism of LAAT. 
	\begin{equation}
		\max_{\|\delta\| \leq \epsilon} L_{\mathrm{CE}}(\textbf{x} + \delta, \textbf{t}_{\mathrm{other}}, \textbf{y})
	\end{equation}
	where \(\textbf{t}_{\mathrm{other}}\) denotes the modified text prompts.

\textbf{Log-Sum-Exp-based} attacks optimize a Log-Sum-Exp (LSE) margin objective over text prompts. This objective serves as a smooth approximation of max-margin loss and adaptively emphasizes the most confusing non-target prompts. Compared to traditional Sum Loss~\cite{weston1998multi}, which uniformly penalizes all non-target classes, the LSE formulation focuses on the most confusing text prompts, thereby providing a stronger and more stable surrogate for worst-case optimization. This formulation mimics the potency of the C\&W loss~\cite{cw} while maintaining gradient stability through the temperature parameter $\tau$, leading to a more reliable assessment of adversarial consistency~\cite{awasthi2023theoretically}. The attack is formulated as:
\begin{equation}
	\max_{|\delta| \leq \epsilon} \Big(\tau \cdot \log \sum_{n \neq m} \exp(\frac{f(\textbf{x} + \delta) \cdot g(\textbf{t}_n)^\top}{\tau}) - f(\textbf{x} + \delta) \cdot g(\textbf{t}_m)^\top \Big)
\end{equation}
where \(\delta\) denotes the adversarial perturbation bounded by \(\epsilon\), \(f(\textbf{x} + \delta)\) denotes the embedding of the adversarial example \(\textbf{x} + \delta\), \(g(\textbf{t})\) denotes the embedding of the text prompt \(\textbf{t}\), \(m\) denotes the ground-truth class and \(n \neq m\) denotes non-target classes. The temperature parameter \(\tau\) controls the sharpness of the soft margin.

To adhere more closely to the adaptive attack principle, we extend the attack objective by augmenting the base LSE loss with method-specific terms that directly target the defense mechanisms of each baseline.

\noindent $\bullet$ \textbf{FT-Adv.} fine-tunes CLIP on adversarial examples using the original contrastive loss. During evaluation, adversarial examples are generated by directly optimizing a margin-based loss over text prompts.
\begin{equation}
	\max_{\|\delta\| \leq \epsilon} (\tau \cdot \log \sum_{n \neq m} \exp(\frac{f(\textbf{x} + \delta) \cdot g(\textbf{t}_n)^\top}{\tau}) - f(\textbf{x} + \delta) \cdot g(\textbf{t}_m)^\top )
\end{equation}
where \(\delta\) denotes the adversarial perturbation bounded by \(\epsilon\), \(f(\textbf{x} + \delta)\) denotes the embedding of the adversarial example \(\textbf{x} + \delta\), \(g(\textbf{t})\) denotes the embedding of the text prompt \(\textbf{t}\), \(m\) denotes the ground-truth class and \(n \neq m\) denotes non-target classes. The temperature parameter \(\tau\) controls the sharpness of the soft margin.

\noindent $\bullet$ \textbf{FARE}~\cite{FARE} constrains the distance between the original and target model embeddings. During evaluation, adversarial examples are generated by directly optimizing a margin-based loss over text prompts and an instance-level feature deviation. The combined loss ensures the attack aligns with both the classification and feature stability objectives of the defended model.
\begin{equation}
	\begin{aligned}
		\max_{\|\delta\| \leq \epsilon} (&\tau \cdot \log \sum_{n \neq m} \exp(\frac{f(\textbf{x} + \delta) \cdot g(\textbf{t}_n)^\top}{\tau}) - f(\textbf{x} + \delta) \cdot g(\textbf{t}_m)^\top \\
		&+ \lambda \cdot (\tau \cdot \log \sum_{i \neq j} \exp(\frac{f(\textbf{x} + \delta) \cdot f^{\mathrm{ori}}(\textbf{x}_i)^\top}{\tau}) \\
		& - f(\textbf{x} + \delta) \cdot f^{\mathrm{ori}}(\textbf{x}_j)^\top))
	\end{aligned}
\end{equation}
where \(f^{\mathrm{ori}}(\textbf{x}_i)\) denotes the corresponding image embedding extracted by the original frozen model from the clean input \(\textbf{x}_i\). \(i\) denotes the matching original image and \(i \neq j\) denotes other instances in the batch. The temperature parameter \(\tau\) controls the sharpness of the soft margin.

\noindent $\bullet$ \textbf{PMG-AFT}~\cite{PMG-AFT} aligns the features of adversarial and clean examples in the target model by minimizing their KL divergence. During evaluation, adversarial examples are generated by directly optimizing a margin-based loss over text prompts and the KL divergence. The combined loss ensures the attack aligns with both the classification and optimization objectives of the defended model.
\begin{equation}
	\begin{aligned}
		&\max_{\|\delta\| \leq \epsilon} (\tau \cdot \log \sum_{n \neq m} \exp(\frac{f(\textbf{x} + \delta) \cdot g(\textbf{t}_n)^\top}{\tau}) - f(\textbf{x} + \delta) \cdot g(\textbf{t}_m)^\top \\
		& + \lambda \cdot L_{\textrm{KL}}(\mathrm{softmax}(f(\textbf{x} + \delta) \cdot g(\textbf{t})^\top) \| \mathrm{softmax}(f(\textbf{x}) \cdot g(\textbf{t})^\top)))        
	\end{aligned}
\end{equation}
where \(\mathrm{softmax}(\cdot, \cdot)\) denotes the softmax operation.

\noindent $\bullet$ \textbf{TGA-ZSR} improves model adversarial robustness by aligning the attention maps of adversarial examples in the target model with those of clean examples in the original model. During evaluation, adversarial examples are generated by directly optimizing a margin-based loss over text prompts together with an L2 loss that enforces attention-map consistency between the two models.
\begin{equation}
	\begin{aligned}
		\max_{\|\delta\| \leq \epsilon} (\tau &\cdot \log \sum_{n \neq m} \exp(\frac{f(\textbf{x} + \delta) \cdot g(\textbf{t}_n)^\top}{\tau}) - f(\textbf{x} + \delta) \cdot g(\textbf{t}_m)^\top \\
		& + \lambda \cdot \| \mathcal{A}(\mathbf{x}+\delta)_{\textrm{tar}} - \mathcal{A}(\mathbf{x})_{\textrm{ori}} \|_2)   \\
		\text{where,} &\ \mathcal{A}(\mathbf{x}) = f_{\textrm{g}}(\mathbf{x}) \cdot g(\mathbf{t})^ \top
	\end{aligned}
\end{equation}
we feed the adversarial sample \(\textbf{x}+\delta\) into the target model and the clean example \(\textbf{x}\) into the original model, obtaining the adversarial attention map \(\mathcal{A}(\mathbf{x}+\delta)_{\textrm{tar}}\) and the clean attention map \(\mathcal{A}(\mathbf{x})_{\textrm{ori}}\), respectively. Here, $f_{\textrm{g}}(\mathbf{x})$ represents the global image feature before the pooling operation of $f(\mathbf{x})$.

\begin{table*}[htbp]
	\centering
	\caption{Zero-shot robust mIoU on images attacked with PGD~\cite{PGD} \(\varepsilon = 8/255\) and clean mIoU. We performed several different methods on MS COCO~\cite{coco}.}
	\label{tab:object_v2}
	\begin{adjustbox}{width=\textwidth,keepaspectratio}
\begin{tabular}{c|ccccccc >{\columncolor{LightPurple}}c >{\columncolor{LightPurple}}c }
\toprule
Methods &CLIP~\cite{CLIP} &FT-Clean &FT-Adv. &TeCoA~\cite{TeCoA} &FARE~\cite{FARE} &LAAT~\cite{LAAT} &PMG-AFT~\cite{PMG-AFT} &TGA-ZSR (ours) &Comp-TGA (ours) \\

\midrule
Robust &0.2767 &0.3247 &0.3826 &0.4477 &0.2569 &0.4505 &0.4481 &\underline{0.4604} &\textbf{0.4612} \\
Clean &0.3819 &0.3752 &0.3873 &0.4537 &0.3944 &0.4551 &0.4546 &\textbf{0.4691} &\underline{0.4689} \\
Average &0.3293 &0.3500 &0.3850 &0.4507 &0.3257 &0.4528 &0.4514 &\underline{0.4648} &\textbf{0.4651} \\

\bottomrule
\end{tabular}
\end{adjustbox}
\end{table*}

\begin{table*}[htbp]
	\centering
	\caption{Zero-shot robust accuracy on images attacked with \(\varepsilon\) of 1/255 of APGD~\cite{AutoAttack}. The optimal accuracy is highlighted in \textbf{bold}, while the second-best accuracy is \underline{underlined}.
	}
	\vspace{-2mm}
	\label{tab:apgd}
	\begin{adjustbox}{width=\textwidth,keepaspectratio}
		\begin{tabular}{c|c|ccccccccccccccc|c}
			
			\toprule
			& & \multicolumn{15}{c|}{\textbf{Zero-shot datasets}} & \\
			\cmidrule(lr){3-17}
			\textbf{Methods} &\textbf{\makecell[c]{\small\rotatebox{75}{Tiny-ImageNet}}}  & \textbf{\makecell[c]{\small\rotatebox{75}{CIFAR-10}}} & \textbf{\makecell[c]{\small\rotatebox{75}{CIFAR-100}}} & \textbf{\makecell[c]{\small\rotatebox{75}{STL-10}}} & \textbf{\makecell[c]{\small\rotatebox{75}{SUN397}}} & \textbf{\makecell[c]{\small\rotatebox{75}{Food101}}} & \textbf{\makecell[c]{\small\rotatebox{75}{Oxfordpets}}} & \textbf{\makecell[c]{\small\rotatebox{75}{Flowers102}}} & \textbf{\makecell[c]{\small\rotatebox{75}{DTD}}}   & \textbf{\makecell[c]{\small\rotatebox{75}{EuroSAT}}} & \textbf{\makecell[c]{\small\rotatebox{75}{FGVC-Aircraft}}} & \textbf{\makecell[c]{\small\rotatebox{75}{ImageNet}}} & \textbf{\makecell[c]{\small\rotatebox{75}{Caltech-101}}} & \textbf{\makecell[c]{\small\rotatebox{75}{Caltech-256}}} & \textbf{\makecell[c]{\small\rotatebox{75}{StanfordCars}}} & \textbf{\makecell[c]{\small\rotatebox{75}{PCAM}}}  &\(\boldsymbol{A_{\textrm{Robust}}}\) \\
			
			\midrule
			CLIP~\cite{CLIP} &0.02 &0.02 &0.06 &0.40 &0.07 &0.03 &0.00 &0.03 &0.27 &0.06 &0.06 &0.51 &0.73 &0.32 &3.89 &0.06 &0.41  \\ 
			
			FT-Clean &0.15 &0.02 &0.01 &1.79 &0.22 &0.04 &0.03 &0.07 &0.75 &0.01 &0.03 &0.50 &2.59 &1.49 &2.41 &0.01 &0.63  \\  
			
			FT-Adv. &52.46 &39.63 &21.57 &69.53 &17.27 &11.76 &34.86 &18.95 &15.21 &11.86 &1.80 &17.22 &50.87 &40.16 &8.38 &\textbf{48.79} &28.77  \\  
			
			TeCoA~\cite{TeCoA} &36.71 &31.88 &17.25 &66.56 &18.64 &13.52 &36.50 &21.30 &15.05 &11.59 &1.68 &17.25 &55.41 &41.12 &8.81 &47.88 &27.57  \\ 
			
			FARE~\cite{FARE} &28.59 &23.28 &13.55 &60.69 &9.73 &13.93 &27.80 &15.56 &9.20 &0.25 &1.08 &12.18 &47.52 &36.64 &8.86 &3.43 &19.52  \\ 
			
			LAAT~\cite{LAAT} &\underline{53.49} &36.67 &22.05 &67.81 &18.59 &12.51 &36.00 &18.70 &11.22 &\textbf{12.79} &1.53 &18.44 &56.05 &41.65 &7.28 &37.46 &28.27  \\ 
			
			PMG-AFT~\cite{PMG-AFT} &45.20 &\textbf{45.85} &\underline{24.73} &\textbf{74.28} &20.34 &\textbf{17.58} &40.31 &21.21 &13.51 &12.13 &1.71 &19.86 &\textbf{61.28} &44.79 &11.06 &\underline{48.49} &31.40 \\ 
			
			\rowcolor{LightPurple}
			TGA-ZSR (ours) &52.09 &43.50 &23.97 &72.35 &\underline{21.83} &\underline{16.57} &\underline{41.13} &\underline{21.83} &\underline{17.23} &11.36 &\underline{3.12} &\underline{20.56} &58.09 &\textbf{46.98} &\textbf{12.71} &46.46 &\underline{31.86}  \\ 
			
			\rowcolor{LightPurple}
			Comp-TGA (ours) &\textbf{54.41} &\underline{44.60} &\textbf{24.90} &\underline{73.09} &\textbf{21.94} &16.05 &\textbf{41.91} &\textbf{21.96} &\textbf{18.13} &\underline{12.57} &\textbf{3.97} &\textbf{20.71} &\underline{58.21} &\underline{46.32} &\underline{12.02} &48.35 &\textbf{32.45}  \\ 
			
			\bottomrule
		\end{tabular}
	\end{adjustbox}
\end{table*}

\noindent $\bullet$ \textbf{Comp-TGA} improves model adversarial robustness by aligning the complementary attention maps of adversarial examples in the target model with those of clean examples in the original model. During evaluation, adversarial examples are generated by directly optimizing a margin-based loss over text prompts together with an L2 loss that enforces attention-map consistency between the two models.
\begin{equation}
	\begin{aligned}
		\max_{\|\delta\| \leq \epsilon} (\tau &\cdot \log \sum_{n \neq m} \exp(\frac{f(\textbf{x} + \delta) \cdot g(\textbf{t}_n)^\top}{\tau}) - f(\textbf{x} + \delta) \cdot g(\textbf{t}_m)^\top  \\
		&+ \lambda \cdot \| \mathcal{A}(\mathbf{x}+\delta)_{\textrm{tar}} - \mathcal{A}(\mathbf{x})_{\textrm{ori}} \|_2)   \\
		\text{where,} &\ \mathcal{A}(\mathbf{x}) = A(\mathbf{x}) \odot (\mathbf{1}-A_{\textrm{non}}(\mathbf{x})) \\
		&\quad \ \ \quad= (f_{\textrm{g}}(\mathbf{x})\cdot g(\mathbf{t})^ \top) \odot (\mathbf{1}- (f_{\textrm{g}}(\mathbf{x})\cdot g(\mathbf{t}_{\textrm{non}})^ \top))
	\end{aligned}
\end{equation}
where \(\mathbf{t}_{\textrm{non}}\) represents the non-class prompts and \(g(\mathbf{t}_{\textrm{non}})\) denotes the text embeddings of \(\mathbf{t}_{\textrm{non}}\). Here, the attention map denotes a complementary attention map that integrates two forms of foreground attention: attention guided by the class prompt and reversed attention driven by the non-class prompt.

The related experiments are presented in Section IV-C of the main paper. As part of this initial exploration, the adaptive attacks we initially design for each method may require further refinement. Nevertheless, this exploration provides a clear signal that adaptive evaluation is crucial for faithfully assessing zero-shot adversarial robustness. The experimental design also points to an important future direction of evaluating zero-shot adversarial robustness under genuine worst-case conditions rather than relying on simple attacks that may lead to a false sense of robustness. We highlight this issue in the Limitations section, aiming to offer insights that may inform future method design and evaluation protocols and promote more reliable robustness research in the field.

\section{Experiments on Object Detection}
We further examine its robustness in these broader contexts. Nevertheless, since CLIP~\cite{CLIP} was not inherently designed for dense or multimodal tasks, adapting it to such scenarios typically requires integration with additional task-specific architectures (e.g., detectors or decoders). In line with this, we extended our analysis to object detection by coupling CLIP with corresponding models. Importantly, since our goal is to evaluate the adversarial robustness of the proposed method on other tasks, we adopt a zero-shot setting for the downstream evaluations.

\begin{table*}[t]
\centering
\caption{Zero-shot robust accuracy on images attacked with $\varepsilon$ of 1/255, 2/255 and 4/255 of PGD~\cite{PGD}. We performed several different methods on Tiny-ImageNet, using $\varepsilon$ of 1/255 of PGD to generate adversarial examples for training, and evaluated them across 16 datasets. We represent the \textbf{average} accuracy across various attack strengths.}
\label{tab:attack_strength} 
\begin{adjustbox}{width=\textwidth,keepaspectratio}
\begin{tabular}{c|c|ccccccccccccccc|c}

\toprule
& & \multicolumn{15}{c|}{\textbf{Zero-shot datasets}} & \\
\cline{3-17} 
\textbf{Methods} &\textbf{\makecell[c]{\small\rotatebox{75}{Tiny-ImageNet}}}  & \textbf{\makecell[c]{\small\rotatebox{75}{CIFAR-10}}} & \textbf{\makecell[c]{\small\rotatebox{75}{CIFAR-100}}} & \textbf{\makecell[c]{\small\rotatebox{75}{STL-10}}} & \textbf{\makecell[c]{\small\rotatebox{75}{SUN397}}} & \textbf{\makecell[c]{\small\rotatebox{75}{Food101}}} & \textbf{\makecell[c]{\small\rotatebox{75}{Oxfordpets}}} & \textbf{\makecell[c]{\small\rotatebox{75}{Flowers102}}} & \textbf{\makecell[c]{\small\rotatebox{75}{DTD}}}   & \textbf{\makecell[c]{\small\rotatebox{75}{EuroSAT}}} & \textbf{\makecell[c]{\small\rotatebox{75}{FGVC-Aircraft}}} & \textbf{\makecell[c]{\small\rotatebox{75}{ImageNet}}} & \textbf{\makecell[c]{\small\rotatebox{75}{Caltech-101}}} & \textbf{\makecell[c]{\small\rotatebox{75}{Caltech-256}}} & \textbf{\makecell[c]{\small\rotatebox{75}{StanfordCars}}} & \textbf{\makecell[c]{\small\rotatebox{75}{PCAM}}}  &\(\boldsymbol{A_{\textrm{Robust}}}\) \\

\midrule
CLIP~\cite{CLIP} &0.64 &2.15 &0.12 &20.35 &0.52 &5.94 &2.97 &0.72 &0.71 &0.03 &0.00 &0.71 &14.28 &9.18 &0.11 &0.04 &3.65 \\ 

FT-Clean &12.44 &18.80 &4.65 &37.16 &0.43 &0.52 &2.03 &0.41 &0.92 &0.02 &0.01 &0.54 &13.02 &7.96 &0.03 &0.44 &6.21 \\ 

FT-Adv. &29.33 &18.10 &11.06 &45.13 &8.58 &5.65 &16.45 &10.15 &9.72 &9.82 &0.83 &8.81 &33.43 &24.14 &3.80 &\textbf{38.06} &17.07 \\ 

TeCoA~\cite{TeCoA} &18.17 &12.78 &8.12 &39.87 &8.90 &6.53 &16.61 &11.04 &\underline{10.07} &9.88 &0.63 &8.43 &34.94 &23.92 &3.45 &33.20 &15.41 \\ 

FARE~\cite{FARE} &12.41 &9.09 &4.23 &33.72 &2.98 &4.75 &9.67 &5.52 &4.26 &0.05 &0.28 &3.90 &23.97 &16.95 &1.48 &3.43 &8.54 \\

LAAT~\cite{LAAT} &\underline{33.52} &26.32 &12.87 &\underline{60.35} &9.13 &6.03 &19.22 &9.86 &6.65 &\underline{9.97} &0.63 &9.69 &38.66 &26.87 &2.97 &23.39 &18.51 \\

PMG-AFT~\cite{PMG-AFT} &25.30 &21.71 &13.29 &47.69 &11.42 &9.49 &20.68 &12.86 &9.45 &\textbf{10.65} &0.90 &11.28 &\underline{41.86} &28.38 &5.40 &\underline{37.88} &19.27 \\ 

\rowcolor{LightPurple}
TGA-ZSR (ours) &\makecell[c]{33.18 \\ \scriptsize{($\pm$ 0.24)}} &\makecell[c]{\underline{27.49} \\ \scriptsize{($\pm$ 0.28)}} &\makecell[c]{\underline{14.43} \\ \scriptsize{($\pm$ 0.24)}} &\makecell[c]{56.48 \\ \scriptsize{($\pm$ 0.20)}}  &\makecell[c]{\underline{12.24} \\ \scriptsize{($\pm$ 0.45)}} &\makecell[c]{\underline{12.57} \\ \scriptsize{($\pm$ 0.48)}}  &\makecell[c]{\underline{24.50} \\ \scriptsize{($\pm$ 0.59)}} &\makecell[c]{\underline{13.04} \\ \scriptsize{($\pm$ 0.45)}}  &\makecell[c]{9.51 \\ \scriptsize{($\pm$ 0.57)}}  &\makecell[c]{7.07 \\ \scriptsize{($\pm$ 0.62)}} &\makecell[c]{\underline{1.61} \\ \scriptsize{($\pm$ 0.12)}}  &\makecell[c]{\underline{11.47} \\ \scriptsize{($\pm$ 0.34)}} &\makecell[c]{40.84 \\ \scriptsize{($\pm$ 0.70)}}  &\makecell[c]{\underline{32.37} \\ \scriptsize{($\pm$ 0.14)}} &\makecell[c]{\underline{6.73} \\ \scriptsize{($\pm$ 0.55)}}  &\makecell[c]{18.98 \\ \scriptsize{($\pm$ 1.21)}} &\makebox[2pt][c]{\makecell[c]{\underline{20.16} \\ \scriptsize{($\pm$ 0.39)}}} \\

\rowcolor{LightPurple}
Comp-TGA (ours) &\makecell[c]{\textbf{39.54} \\ \scriptsize{($\pm$ 0.57)}} &\makecell[c]{\textbf{37.28} \\ \scriptsize{($\pm$ 0.92)}} &\makecell[c]{\textbf{19.36} \\ \scriptsize{($\pm$ 0.42)}} &\makecell[c]{\textbf{61.80} \\ \scriptsize{($\pm$ 0.59)}}  &\makecell[c]{\textbf{13.63} \\ \scriptsize{($\pm$ 0.45)}} &\makecell[c]{\textbf{13.66} \\ \scriptsize{($\pm$ 0.38)}}  &\makecell[c]{\textbf{28.41} \\ \scriptsize{($\pm$ 0.78)}} &\makecell[c]{\textbf{14.48} \\ \scriptsize{($\pm$ 0.65)}}  &\makecell[c]{\textbf{11.07} \\ \scriptsize{($\pm$ 0.46)}}  &\makecell[c]{8.78 \\ \scriptsize{($\pm$ 0.26)}} &\makecell[c]{\textbf{1.95} \\ \scriptsize{($\pm$ 0.06)}}  &\makecell[c]{\textbf{12.60} \\ \scriptsize{($\pm$ 0.35)}} &\makecell[c]{\textbf{45.71} \\ \scriptsize{($\pm$ 0.48)}}  &\makecell[c]{\textbf{35.08} \\ \scriptsize{($\pm$ 0.43)}} &\makecell[c]{\textbf{7.75} \\ \scriptsize{($\pm$ 0.36)}}  &\makecell[c]{20.21 \\ \scriptsize{($\pm$ 0.87)}} &\makebox[2pt][c]{\makecell[c]{\textbf{23.21} \\ \scriptsize{($\pm$ 0.36)}}}  \\

\bottomrule
\end{tabular}
\end{adjustbox}
% \vspace{-4mm}
\end{table*}
\begin{table*}[htbp]
	\centering
	\caption{Zero-shot robust accuracy on images attacked with \(\varepsilon\) of 8/255 of PGD~\cite{PGD}. The optimal accuracy is highlighted in \textbf{bold}, while the second-best accuracy is \underline{underlined}.}
	\label{tab:8_255}
	\begin{adjustbox}{width=\textwidth,keepaspectratio}
		\begin{tabular}{c|c|ccccccccccccccc|c}
			
			\toprule
			& & \multicolumn{15}{c|}{\textbf{Zero-shot datasets}} & \\ 
			\cmidrule(lr){3-17}
			\textbf{Methods} &\textbf{\makecell[c]{\small\rotatebox{75}{Tiny-ImageNet}}}  & \textbf{\makecell[c]{\small\rotatebox{75}{CIFAR-10}}} & \textbf{\makecell[c]{\small\rotatebox{75}{CIFAR-100}}} & \textbf{\makecell[c]{\small\rotatebox{75}{STL-10}}} & \textbf{\makecell[c]{\small\rotatebox{75}{SUN397}}} & \textbf{\makecell[c]{\small\rotatebox{75}{Food101}}} & \textbf{\makecell[c]{\small\rotatebox{75}{Oxfordpets}}} & \textbf{\makecell[c]{\small\rotatebox{75}{Flowers102}}} & \textbf{\makecell[c]{\small\rotatebox{75}{DTD}}}   & \textbf{\makecell[c]{\small\rotatebox{75}{EuroSAT}}} & \textbf{\makecell[c]{\small\rotatebox{75}{FGVC-Aircraft}}} & \textbf{\makecell[c]{\small\rotatebox{75}{ImageNet}}} & \textbf{\makecell[c]{\small\rotatebox{75}{Caltech-101}}} & \textbf{\makecell[c]{\small\rotatebox{75}{Caltech-256}}} & \textbf{\makecell[c]{\small\rotatebox{75}{StanfordCars}}} & \textbf{\makecell[c]{\small\rotatebox{75}{PCAM}}}  &\(\boldsymbol{A_{\textrm{Robust}}}\) \\
			
			\midrule
			CLIP~\cite{CLIP} &0.22 &1.55 &0.02 &14.03 &0.22 &\textbf{4.88} &1.80 &0.34 &0.00 &0.00 &0.00 &0.36 &9.29 &5.83 &\underline{0.09} &0.00 &2.42  \\ 
			
			FT-Clean &10.76 &\underline{16.20} &4.38 &34.40 &0.19 &0.40 &1.72 &0.21 &0.00 &\underline{0.03} &0.00 &0.23 &11.66 &6.53 &0.00 &0.00 &5.42  \\ 
			
			FT-Adv. &5.56 &1.53 &1.32 &13.48 &0.06 &0.02 &0.57 &0.11 &0.48 &0.00 &0.00 &0.11 &8.43 &3.43 &0.00 &\underline{0.16} &2.20  \\ 
			
			TeCoA~\cite{TeCoA} &1.56 &0.32 &0.21 &9.46 &0.06 &0.10 &0.93 &0.15 &0.21 &0.00 &0.00 &0.17 &9.34 &3.94 &0.01 &0.04 &1.66  \\ 
			
			FARE~\cite{FARE} &5.14 &2.26 &0.61 &15.36 &0.14 &1.33 &1.85 &0.20 &0.00 &0.00 &0.00 &0.23 &9.28 &5.74 &0.03 &0.00 &2.64  \\ 
			
			LAAT~\cite{LAAT} &\underline{15.06} &\textbf{20.55} &\textbf{6.25} &\textbf{55.33} &0.66 &0.80 &5.67 &0.24 &\textbf{0.96} &\textbf{0.04} &0.00 &1.11 &15.32 &10.90 &0.03 &\textbf{2.93} &\underline{8.49} \\ 
			
			PMG-AFT~\cite{PMG-AFT} &0.43 &0.15 &0.04 &3.98 &0.03 &0.02 &0.00 &0.03 &\underline{0.80} &0.00 &0.00 &0.05 &5.09 &1.63 &0.01 &0.09 &0.82 \\
			
			\rowcolor{LightPurple}
			TGA-ZSR (ours) &13.13 &5.76 &2.49 &33.59 &\underline{0.74} &1.07 &\underline{5.86} &\underline{0.98} &0.21 &0.00 &0.00 &\underline{1.59} &\underline{17.41} &\underline{12.34} &0.05 &0.00 &5.95  \\ 
			
			\rowcolor{LightPurple}
			Comp-TGA (ours) &\textbf{20.56} &14.17 &\underline{5.45} &\underline{41.28} &\textbf{1.25} &\underline{1.98} &\textbf{9.24} &\textbf{1.40} &0.48 &0.00 &0.00 &\textbf{2.39} &\textbf{23.25} &\textbf{16.20} &\textbf{0.10} &0.00 &\textbf{8.61} \\ 
			
			\bottomrule
		\end{tabular}
	\end{adjustbox}
\end{table*}

Inspired by ViLD~\cite{vild}, we integrate CLIP with Faster R-CNN~\cite{ren2016faster} to establish a zero-shot object detection framework. Specifically, Faster R-CNN is first used to generate candidate bounding boxes for each image. The corresponding regions are then cropped and individually fed into CLIP, which predicts the object category by selecting the crop whose visual feature is most similar to the text embedding of each class.

To evaluate CLIP's robustness, we apply PGD~\cite{PGD} attack with \(\epsilon = 8/255\) on the cropped images. The performance is evaluated on the 2017 version of the MS COCO~\cite{coco} dataset, which contains 5000 test images across 80 object categories, using the IoU metric. As shown in Table~\ref{tab:object_v2}, our proposed methods, TGA-ZSR and Comp-TGA, achieve the best performance across all metrics. Specifically, Comp-TGA attains a robust mIoU of 0.4612 and a clean mIoU of 0.4689, outperforming prior approaches such as LAAT~\cite{LAAT} and PMG-AFT~\cite{PMG-AFT}. These results demonstrate that our approach effectively balances zero-shot robustness and generalization.

\section{Additional Attack Type}
For a more comprehensive and rigorous assessment, we additionally evaluated the models under the adaptive gradient-based APGD~\cite{deepfool} attack as shown in Table~\ref{tab:apgd}. These experiments provide a broader perspective on the model's robustness across different attack types, allowing us to better understand both the strengths and potential vulnerabilities of the fine-tuned models. Overall, this multi-faceted evaluation strategy ensures that our robustness claims are not limited to specific attack configurations and offers a more complete characterization of model behavior under adversarial conditions.

\begin{table*}[htbp]
	\centering
	\caption{Zero-shot robust under AutoAttack~\cite{AutoAttack} and clean accuracy across 16 datasets. The optimal accuracy is highlighted in \textbf{bold}, while the second-best accuracy is \underline{underlined}.}
	\vspace{-2mm}
	\label{tab:aa}  
	\begin{adjustbox}{width=\textwidth,keepaspectratio}
		\begin{tabular}{c|c|c|ccccccccccccccc|c}
			
			\toprule
			& & & \multicolumn{15}{c|}{\textbf{Zero-shot datasets}} & \\
			\cmidrule(lr){4-18}
			\textbf{Test} &\textbf{Methods} &\textbf{\makecell[c]{\small\rotatebox{75}{Tiny-ImageNet}}}  & \textbf{\makecell[c]{\small\rotatebox{75}{CIFAR-10}}} & \textbf{\makecell[c]{\small\rotatebox{75}{CIFAR-100}}} & \textbf{\makecell[c]{\small\rotatebox{75}{STL-10}}} & \textbf{\makecell[c]{\small\rotatebox{75}{SUN397}}} & \textbf{\makecell[c]{\small\rotatebox{75}{Food101}}} & \textbf{\makecell[c]{\small\rotatebox{75}{Oxfordpets}}} & \textbf{\makecell[c]{\small\rotatebox{75}{Flowers102}}} & \textbf{\makecell[c]{\small\rotatebox{75}{DTD}}}   & \textbf{\makecell[c]{\small\rotatebox{75}{EuroSAT}}} & \textbf{\makecell[c]{\small\rotatebox{75}{FGVC-Aircraft}}} & \textbf{\makecell[c]{\small\rotatebox{75}{ImageNet}}} & \textbf{\makecell[c]{\small\rotatebox{75}{Caltech-101}}} & \textbf{\makecell[c]{\small\rotatebox{75}{Caltech-256}}} & \textbf{\makecell[c]{\small\rotatebox{75}{StanfordCars}}} & \textbf{\makecell[c]{\small\rotatebox{75}{PCAM}}}  &Average \\
			
			\midrule
			\multirow{3}{*}{Robust} &PMG-AFT~\cite{PMG-AFT} &44.26 &\textbf{44.12} &\textbf{23.66} &\textbf{73.90} &19.63 &\textbf{17.25} &39.25 &20.87 &13.72 &\underline{11.99} &1.68 &19.17 &\textbf{60.57} &44.25 &9.59 &\textbf{48.53} &\underline{30.78} \\
			
			&TGA-ZSR (ours) &\underline{49.49} &40.69 &22.30 &71.97 &\underline{20.14} &\underline{15.41} &\underline{39.82} &\underline{21.35} &\underline{16.81} &11.31 &\underline{2.34} &\underline{19.22} &\underline{57.65} &\textbf{45.73} &\underline{10.14} &\underline{48.01} &30.77 \\ 
			
			&Comp-TGA (ours) &\textbf{53.00} &\underline{42.18} &\underline{23.58} &\underline{73.04} &\textbf{20.47} &15.30 &\textbf{40.28} &\textbf{22.37} &\textbf{17.17} &\textbf{12.28} &\textbf{3.43} &\textbf{19.47} &57.58 &\underline{45.11} &\textbf{10.24} &47.05 &\textbf{31.41} \\ 
			
			\midrule
			\multirow{3}{*}{Clean} &PMG-AFT\cite{PMG-AFT} &67.11 &74.62 &44.68 &88.85 &37.42 &\underline{37.47} &66.34 &35.66 &21.17 &17.76 &4.71 &35.93 &\textbf{76.70} &61.96 &25.21 &\underline{49.99} &46.60 \\ 
			
			&TGA-ZSR (ours) &\underline{76.58} &\textbf{79.24} &\textbf{48.35} &\textbf{90.78} &\textbf{44.27} &\textbf{40.39} &\textbf{69.86} &\textbf{39.73} &\underline{26.22} &\textbf{19.22} &\textbf{8.88} &\textbf{40.08} &\underline{76.17} &\textbf{67.14} &\textbf{29.96} &49.90 &\textbf{50.42} \\ 
			
			&Comp-TGA (ours) &\textbf{76.82} &\underline{76.36} &\underline{45.88} &\underline{90.30} &\underline{42.24} &36.19 &\underline{67.94} &\underline{38.78} &\textbf{26.64} &\underline{18.63} &\underline{8.35} &\underline{37.97} &75.82 &\underline{64.81} &\underline{25.54} &\textbf{50.73} &\underline{48.94} \\ 
			
			\bottomrule
		\end{tabular}
	\end{adjustbox}
\end{table*}

\begin{table*}[htbp]
	\centering
	\caption{Zero-shot robust accuracy under PGD~\cite{PGD} and clean accuracy across 16 datasets on TGA-ZSR. We performed several different hyperparameters on Tiny-ImageNet and evaluated across 16 datasets. 
	}
	\label{tab:larger_param_tga}
	\subfloat[\textrm{\small Zero-shot robust accuracy on TGA-ZSR.}]{
		\centering
		\begin{adjustbox}{width=0.49\textwidth,keepaspectratio}
			\tiny
			\begin{tabular}{c|cccc}
				\toprule
				\diagbox{\(\alpha\)}{\(\beta\)} & 0 &\(10^{-3}\) & \(10^{-2}\) & \(10^{-1}\) \\
				\midrule
				0     &29.45   &29.54   &30.25   &26.72   \\
				\(10^{-3}\)  &29.55   &29.61   &30.30   &26.71   \\
				\(10^{-2}\)  &30.24   &30.32   &31.12   &27.44   \\
				\(10^{-1}\)  &31.44   &31.59   &31.43   &31.10   \\
				\bottomrule
			\end{tabular}
		\end{adjustbox}
		\label{tab:larger_param_tga_adv}
	}
	\subfloat[\textrm{\small Zero-shot clean accuracy on TGA-ZSR.}]{
		\centering
		\begin{adjustbox}{width=0.49\textwidth,keepaspectratio}
			\tiny
			\begin{tabular}{c|cccc}
				\toprule
				\diagbox{\(\alpha\)}{\(\beta\)} & 0 &\(10^{-3}\) & \(10^{-2}\) & \(10^{-1}\) \\
				\midrule
				0     &44.97   &45.06   &46.08   &58.98   \\
				\(10^{-3}\)  &45.09   &45.17   &46.16   &58.84   \\
				\(10^{-2}\)  &46.18   &46.27   &47.02   &59.18   \\
				\(10^{-1}\)  &50.42   &40.52   &50.69   &56.26   \\
				\bottomrule
			\end{tabular}
		\end{adjustbox}
		\label{tab:larger_param_tga_cle}
	}
\end{table*}

\begin{table*}[htbp]
	\centering
	\caption{Zero-shot robust accuracy under PGD~\cite{PGD} and clean accuracy across 16 datasets on Comp-TGA. We performed several different hyperparameters on Tiny-ImageNet and evaluated across 16 datasets. 
	}
	\label{tab:larger_param_comp}
	\subfloat[\textrm{\small Zero-shot robust accuracy on Comp-TGA.}]{
		\centering
		\begin{adjustbox}{width=0.49\textwidth,keepaspectratio}
			\tiny
			\begin{tabular}{c|cccc}
				\toprule
				\diagbox{\(\alpha\)}{\(\beta\)} & 0 &\(10^{-3}\) & \(10^{-2}\) & \(10^{-1}\) \\
				\midrule
				0     &29.45   &29.43   &29.58   &42.69   \\
				\(10^{-3}\)  &29.50   &29.53   &29.58   &42.65   \\
				\(10^{-2}\)  &29.53   &29.57   &29.67   &42.74   \\
				\(10^{-1}\)  &30.36   &30.37   &29.84   &44.46   \\
				\bottomrule
			\end{tabular}
		\end{adjustbox}
		\label{tab:larger_param_comp_adv}
	}
	\subfloat[\textrm{\small Zero-shot clean accuracy on Comp-TGA.}]{
		\centering
		\begin{adjustbox}{width=0.49\textwidth,keepaspectratio}
			\tiny
			\begin{tabular}{c|cccc}
				\toprule
				\diagbox{\(\alpha\)}{\(\beta\)} & 0 &\(10^{-3}\) & \(10^{-2}\) & \(10^{-1}\) \\
				\midrule
				0     &44.97   &45.01   &45.22   &55.19   \\
				\(10^{-3}\)  &44.96   &45.02   &45.20   &55.21   \\
				\(10^{-2}\)  &45.21   &45.25   &45.33   &55.26   \\
				\(10^{-1}\)  &46.16   &46.25   &46.33   &55.44   \\
				\bottomrule
			\end{tabular}
		\end{adjustbox}
		\label{tab:larger_param_comp_cle}
	}
	\vspace{-3mm}
\end{table*}

\section{Experiments with Different Attack Strengths} 
Table~\ref{tab:attack_strength} reports robust accuracy inder moderate perturbations (\(\epsilon =\) 1/255, 2/255, and 4/255) while Table~\ref{tab:8_255} focuses on stronger perturbations (\(\epsilon =\) 8/255). By merging them, Table VIII provides a comprehensive view of model performance across the full range of attack strengths. The results show that all methods exhibit decreasing accuracy as \(\epsilon\) increases. LAAT~\cite{LAAT} enhances the differences between category text embeddings through an expansion algorithm, while PMG-AFT~\cite{PMG-AFT} relies on a pre-trained model for guided adversarial fine-tuning. Although both methods achieve some improvements on certain tasks, they incur high computational costs (PMG-AFT memory usage is 18449 Mb compared to our 16265 Mb and 16293 Mb) or long training times (LAAT training time per batch is 3.47 seconds compared to our 0.96 and 1.10 seconds), and they fail to maintain zero-shot generalization and robustness across different attacks and tasks. Their zero-shot generalization (clean accuracy) is typically 8 to 10 points lower than ours, and they struggle to maintain robustness under stronger attacks such as CW, AutoAttack, and APGD, as well as cross-task benchmarks like object detection. In contrast, our methods underperform under some high perturbation levels, but they achieve the highest average robust accuracy with lower computational cost and demonstrate stronger cross-scenario adaptability and scalability.

\begin{figure*}[t]
	\centering
	\subfloat[\textrm{\small Zero-shot robust accuracy.}]{
		\includegraphics[width=0.32\textwidth]{figures/TGA_robust.png} 
		\label{fig:tga_param_robust}
	}
	\subfloat[\textrm{\small Zero-shot clean accuracy.}]{
		\includegraphics[width=0.32\textwidth]{figures/TGA_clean.png}  
		\label{fig:tga_param_clean}
	}
	\subfloat[\textrm{\small Zero-shot overall accuracy.}]{
		\includegraphics[width=0.32\textwidth]{figures/TGA_overall.png}  
		\label{fig:tga_param_overall}
	}
	\caption{Zero-shot robust accuracy under PGD~\cite{PGD} and clean accuracy across 16 datasets for TGA-ZSR. Several different hyperparameter settings were explored on Tiny-ImageNet and evaluated across 16 datasets.}
	\label{fig:tga_param}
	\vspace{-3mm}
\end{figure*}

\begin{figure*}[t]
	\centering
	\subfloat[\textrm{\small Zero-shot robust accuracy.}]{
		\includegraphics[width=0.32\textwidth]{figures/Comp_robust.png} 
		\label{fig:comp_param_robust}
	}
	\subfloat[\textrm{\small Zero-shot clean accuracy.}]{
		\includegraphics[width=0.32\textwidth]{figures/Comp_clean.png}  
		\label{fig:comp_param_clean}
	}
	\subfloat[\textrm{\small Zero-shot overall accuracy.}]{
		\includegraphics[width=0.32\textwidth]{figures/Comp_overall.png}  
		\label{fig:comp_param_overall}
	}
	\caption{Zero-shot robust accuracy under PGD~\cite{PGD} and clean accuracy across 16 datasets for Comp-TGA. Several different hyperparameter settings were explored on Tiny-ImageNet and evaluated across 16 datasets.}
	\label{fig:comp_param}
\end{figure*}

\section{Clean Accuracy under AutoAttack Setting}
We randomly split the Tiny-ImageNet training set, using 80\% for training and the remaining 20\% for validation to select the optimal hyperparameters. The model was trained on the entire training set using adversarial examples generated by PGD with $\varepsilon=1/255$, and validated under PGD with the same attack strength. Optimal hyperparameters ($\alpha$=0.08, $\beta$=0.009 for TGA-ZSR and $\alpha$=0.10, $\beta$=0.07 for Comp-TGA) were selected based on validation performance. We further provide both robust and clean accuracy under AutoAttack~\cite{AutoAttack} in Table~\ref{tab:aa} to offer a more complete view of the model's performance. The results show that both clean and robust accuracy remain strong. Adjusting the parameters to AutoAttack still produces solid results, with clean accuracy remaining higher than PMG-AFT, suggesting that the method maintains generalization while strengthening robustness.

\section{The Analysis of Hyperparameter effects}
We extended our hyperparameter experiments by evaluating \(\alpha,\beta \in \{0, 10^{-3},10^{-2},10^{-1}\}\) on both TGA-ZSR and Comp-TGA. The results in Table~\ref{tab:larger_param_tga} and~\ref{tab:larger_param_comp} reveal consistent trends across both TGA-ZSR and Comp-TGA. As \(\alpha\) and \(\beta\) increase, moderate weighting leads to simultaneous improvements in both robust and clean accuracies. However, excessively large values result in performance degradation.

Building upon the ablation results presented in Table X of the paper and the extended ablation results in Table~\ref{tab:larger_param_tga} and~\ref{tab:larger_param_comp}, we further visualize the influence of \(\alpha\) and \(\beta\) as contour heatmaps in Fig.~\ref{fig:tga_param} and Fig.~\ref{fig:comp_param}, where the x-axis and y-axis correspond to \(\alpha\) and \(\beta\), respectively, and the color indicates the averaged performance over all datasets. These visualizations clearly illustrate the impact of \(\alpha\) and \(\beta\) on model performance. They also reveal the sensitivity of our methods to these hyperparameters, and Comp-TGA shows lower sensitivity than TGA-ZSR, which further supports the advantages of our improved design.

\begin{table}[t]
	\centering
	\caption{Ablation study on Hyper-parameters. We trained several different methods on Tiny-ImageNet using PGD with $\varepsilon = 1/255$, and evaluated them across 16 datasets under PGD with the same attack strength.} 
	{\small
		\label{tab:hyper}  
		\begin{tabular}{p{1.5cm}<{\centering}|p{0.6cm}<{\centering}p{0.6cm}<{\centering}|p{1.1cm}<{\centering}p{1.1cm}<{\centering}p{1.1cm}<{\centering}}
			\toprule
			\textbf{Methods} & \(\boldsymbol{\alpha}\) & \textbf{\(\boldsymbol{\beta}\)} &{\footnotesize \(\boldsymbol{A_{\textrm{Robust}}}\)}  &{\footnotesize \(\boldsymbol{A_{\textrm{Clean}}}\)} &{\footnotesize \(\boldsymbol{A_{\textrm{Overall}}}\)}\\ 
			\midrule
			\multirow{5}{*}{TGA-ZSR} &0.07 &0.05 &40.43 &56.86 &48.65 \\ 
			&0.08 &0.04 &37.28 &55.96 &46.62 \\ 
			&0.08 &0.06 &40.16 &\textbf{56.88} &48.52 \\
			&0.09 &0.05 &32.31 &54.63 &43.47 \\
			\rowcolor{LightPurple}
			\cellcolor{white}        &0.08 &0.05 &\textbf{42.09} &56.44 &\textbf{49.27} \\
			\midrule
			\multirow{5}{*}{Comp-TGA} &0.10 &0.09 &44.22 &55.29 &49.76 \\ 
			&0.10 &0.11 &44.06 &\textbf{55.80} &49.93 \\ 
			&0.09 &0.10 &44.11 &55.12 &49.61 \\
			&0.11 &0.10 &43.76 &55.49 &49.62 \\
			\rowcolor{LightPurple}
			\cellcolor{white}         &0.10 &0.10 &\textbf{44.46} &55.44 &\textbf{49.95} \\
			
			\bottomrule
	\end{tabular}}
\end{table}
\begin{figure}[t]
	\centering
	\includegraphics[width=0.49\textwidth]{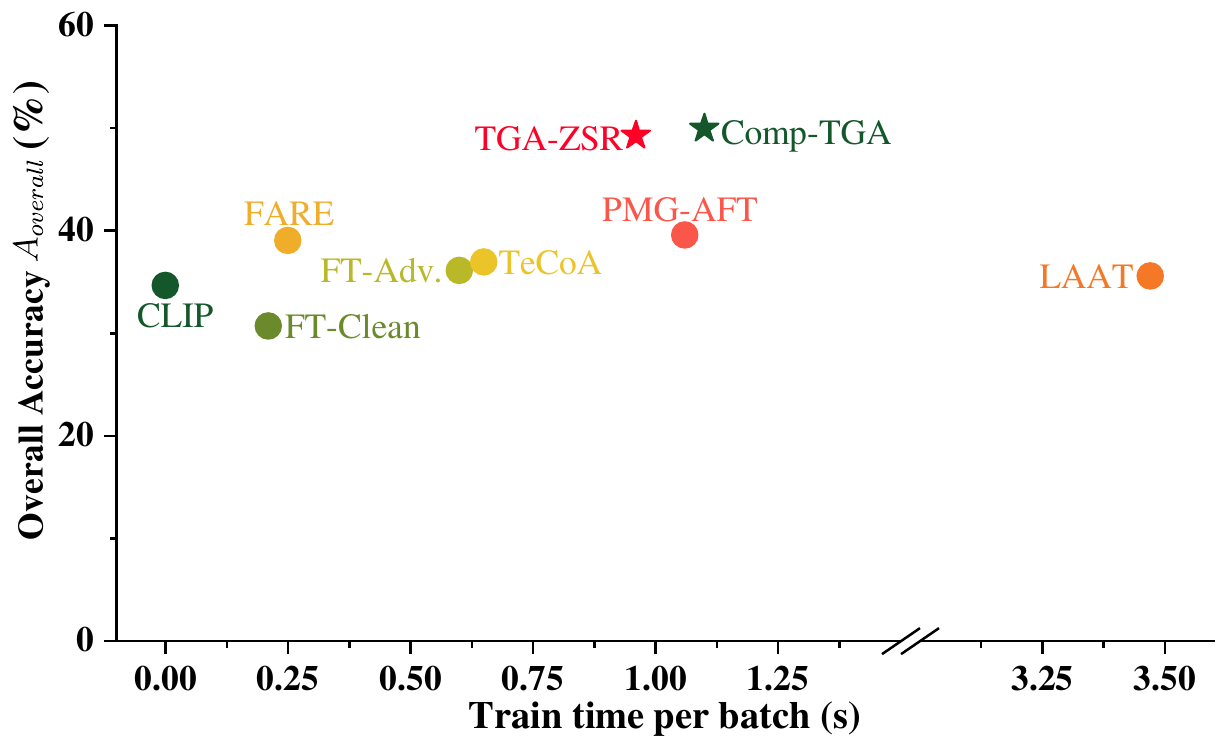}
	\caption{The trade-off between train time and overall accuracy. Each point represents a method, with \ding{108} indicating the baseline models and \ding{72} marking our methods to provide a clear visual distinction.} 
	\label{fig:time_overall}
\end{figure}

\section{Effect of Hyper-parameters}
Following the protocol of previous works (TeCoA~\cite{TeCoA}, PMG-AFT~\cite{PMG-AFT}, FARE~\cite{FARE}), we fine-tuned the CLIP model on adversarial samples from a single dataset (Tiny-ImageNet in our case) for `adversarial fine-tuning' and subsequently evaluated its performance across 15 datasets, including Tiny-ImageNet itself. Thus, we only need to tune hyperparameters on just the training dataset. We randomly selected 80\% of the training set for training and the remaining 20\% for validation to choose the hyperparameters. The model was trained on the entire training set using adversarial examples generated by PGD with $\varepsilon=1/255$, and validated under PGD with the same attack strength. Optimal hyperparameters ($\alpha$=0.08, $\beta$=0.05 for TGA-ZSR and $\alpha$=0.10, $\beta$=0.10 for Comp-TGA) were selected based on validation performance. After determining the optimal values on the Tiny-ImageNet validation set, we further analyze the sensitivity of these hyperparameters by applying the same configurations across 16 datasets in total. Table~\ref{tab:hyper} reports the performance of TGA-ZSR and Comp-TGA on 16 datasets under different hyper-parameter settings. The highlighted rows indicate the selected optimal hyperparameters, which consistently achieve the best or near-best results across these datasets.

\section{The trade-off between train time and overall accuracy}
In our experiments in Fig.~\ref{fig:time_overall} comparing efficiency and overall accuracy across different approaches, FARE~\cite{FARE} achieves high efficiency by relying on a lightweight fine-tuning strategy that skips text processing and dual-branch structures. By contrast, TGA-ZSR and Comp-TGA leverage text-guided attention and a dual-branch architecture to enhance robustness while preserving generalization. Although this design incurs higher training costs compared to FARE, it delivers substantially higher overall accuracy. The advantage becomes even clearer under stronger attacks, where FARE's performance drops more sharply. These observations demonstrate that the modest increase in training time for our methods directly translates into a superior balance between robustness and generalization. In particular, Comp-TGA further refines this balance through improved attention fusion, highlighting its effectiveness as a robust and efficient solution.

	\vfill
	
\end{document}